\documentclass{article} 
\usepackage{iclr2027_conference,times}

\usepackage{amsmath,amsfonts,bm}

\def\eqref#1{equation~\ref{#1}}

\def\1{\bm{1}}

\DeclareMathAlphabet{\mathsfit}{\encodingdefault}{\sfdefault}{m}{sl}
\SetMathAlphabet{\mathsfit}{bold}{\encodingdefault}{\sfdefault}{bx}{n}

\usepackage{hyperref}
\usepackage{url}
\usepackage{booktabs}
\usepackage{tabularx}
\usepackage{xcolor}
\usepackage{colortbl}
\usepackage{hyperref}
\usepackage{graphicx}
\usepackage{wrapfig}
\usepackage{lipsum}
\usepackage{siunitx}
\usepackage{makecell}
\usepackage{tablefootnote}

\usepackage{float}
\usepackage{authblk}

\DeclareSIUnit\baud{baud}
\definecolor{rowhighlight}{HTML}{EAF2F8}  \title{Devol-ONE: One Autoregressive Mixture of Transformers to Unify Vision-Language-Action and Latent World Modeling}

\iclrfinalcopy

\author[1]{Hongyi Cai}
\author[1]{Yi Herng Ong}
\author[1]{Tingshiuan C. Wu}
\author[1]{Chiew Hui Lim}
\author[1]{Hanxia Li}
\author[1]{Kehong Guo}
\author[1]{Sze Yuan Cheong\thanks{Corresponding author: \texttt{elijahong@devolrobots.ai}}}

\affil[1]{Devol Robots, San Francisco, CA, United States}

\begin{document}
\maketitle

\begin{abstract}
Vision Language Action (VLA) models condition actions directly on current visual and language context, without an explicit account of how the scene evolves under candidate actions. World Action Models (WAM) attempt to address this limitation by predicting future states, but existing designs keep prediction and policy learning architecturally separate, connecting them only through the predicted output, whether through pixel space video generation or a latent forecasting module trained independently of the policy. We present Devol-ONE, a Mixture of Transformers architecture that unifies vision language understanding, latent world dynamics prediction, and action generation within a single autoregressive framework. Instead of encoding vision language tokens once and feeding them to the action expert, Devol-ONE runs autoregressive prediction jointly across a vision language stream and a V-JEPA pretrained dynamics stream, attending to the vision language key-value cache at every layer to forecast future latent states under language guidance. The action expert is in turn shaped continuously by semantic reasoning and predicted physical dynamics rather than by a fixed representation computed in advance. Extensive experiments are conducted on LIBERO, LIBERO-PLUS, RoboTwin2.0 along with real-world evaluation on Flexiv single-arm and dual-arm setups. Ablation studies show the effectiveness of dynamic stream prediction and layer-wise unified attention to validate our model architectural coherency.
\end{abstract}

\begin{figure}[!t]  \centering  \includegraphics[width=1\linewidth]{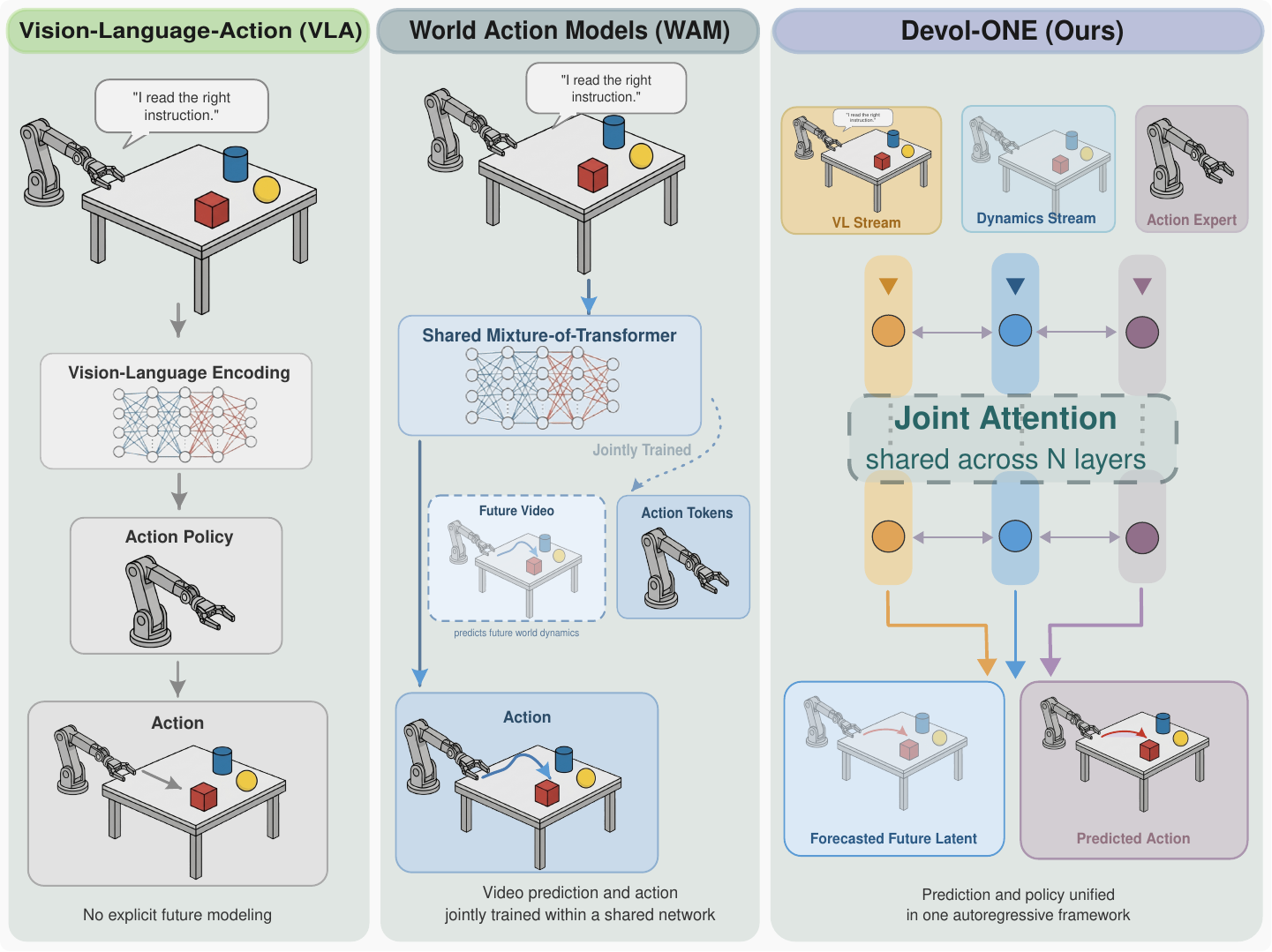}  \caption{Comparison of VLA models, joint WAMs, and Devol-ONE. (\textbf{Left}) Standard VLA models map vision-language features directly to actions. (\textbf{Middle}) Joint WAMs use a shared transformer to predict future video and actions. (\textbf{Right}) Devol-ONE combines a VL stream, a JEPA dynamics stream initialized from a V-JEPA pretrained backbone, and an action expert through layerwise joint attention.}  \label{fig:teaser}
\end{figure}
\section{Introduction}

Imitation learning (IL) is widely used to train robotic manipulation policies. Among existing approaches, Vision-Language-Action (VLA) models \citep{clover24,vpp2024hu,cogact24li,li2024towards,roboflaminggo23li,24pi0,25pi0.5} benefit from large-scale pretraining on visual observations, language prompts, and demonstration trajectories, supporting instruction following across multiple tasks. However, conventional VLAs \citep{lin2025evo0visionlanguageactionmodelimplicit, lin2025evo1lightweightvisionlanguageactionmodel} encode the scene once and map the resulting representation directly to control, without modeling how the environment will evolve in response to an action. 

World-Action Models (WAMs) incorporate predicted future states into action generation \citep{shen2026wamsurvey}. Existing designs can be broadly grouped into two categories. Cascaded designs synthesize future frames or subgoals, then recover actions through a separate inverse-dynamics or goal-conditioned module \citep{du2024learning, du2023video, susie23, clover24, seer24}. Joint designs denoise future video tokens and action tokens within a single backbone \citep{unified25, zhu2025unified, cen2025worldvla, cheang2024gr}. These approaches benefits from internet-scale video generation models that provide transferable dynamics which static image-text pretraining cannot supply and the prediction target that can be supervised with action-free video, allowing the training signal to scale beyond limited action labels.  However, in WAMs, generating future observations in pixel space causes substantial inference costs, as well as lacking explicit text reasoning. On the other hand, even though VLA has its own vision-text semantic space, it does not take real-world physical environment into account.

\textbf{Can an embodied policy jointly reason about its task, predict future world states, and generate actions within a unified architecture?} We present Devol-ONE, a Mixture-of-Transformers (MoT) architecture that integrates latent dynamics prediction with vision-language understanding and action generation. As illustrated in Figure~\ref{fig:teaser}, Devol-ONE decomposes computation into three parameter-disjoint streams: a VL stream providing task-related semantic features, a dynamics stream initialized from a V-JEPA pretrained backbone \citep{assran2025vjepa2} that autoregressively forecasts future latent states $\hat{z}_{t+1},\dots,\hat{z}_{t+N}$ under language guidance, and a diffusion action expert that denoises continuous action chunks. In general, a Layerwise Joint MoT Block connects the streams at every transformer layer, allowing the action expert to attend to both semantic features and predicted dynamics throughout action decoding. 

In VLA-JEPA \citep{vlajepa2026}, a related latent world model, the target latent is constructed from a ground-truth future frame and is therefore available only during training. At inference, the robot has no access to future observations and leave no choices but roll out its own latent predictions, conditioning only on the current state and image. This train--inference mismatch is related to exposure bias in sequence prediction \citep{bengio2015scheduledsamplingsequenceprediction}, as inference relies on model-generated latent states. We introduce Autoregressive Latent Rollout Training (ALRT), which exposes both the predictor and the action expert to model-generated multi-step latent rollouts during training. 

We evaluate Devol-ONE on LIBERO \citep{LIBERO23}, LIBERO-Plus \citep{fei25libero-plus}, and RoboTwin 2.0 \citep{mu2025robotwin}, together with real-world deployment on Flexiv single-arm and dual-arm platforms. Devol-ONE attains 98.4\% average success on LIBERO and 71.4\% on zero-shot LIBERO-Plus without fine-tuning on the perturbed data. On the 34 evaluated RoboTwin 2.0 tasks, the average success rate increases from 61.1\% under the Clean setting to 63.7\% under the Randomized setting. These results suggest improved robustness under the evaluated perturbations. Our contributions are summarized as follows:
\begin{itemize}
    \item \textbf{A unified autoregressive architecture.} Devol-ONE couples a vision-language stream, a V-JEPA latent dynamics stream, and a diffusion action expert through layerwise joint attention, grounding future-latent forecasting and action generation in a shared computation rather than connecting them through a predicted output.
    \item \textbf{Rollout training for the policy, not only the predictor.} To solve exposure bias in latent world models propagates to any action head conditioned on the rollout, and introduce ALRT, which trains both the dynamics stream and the action expert on self-generated $K$-step latent trajectories. 
    \item \textbf{Validation under distribution shift and in the real world.} We evaluate Devol-ONE on simulation benchmarks, including zero-shot LIBERO-Plus and RoboTwin 2.0 under randomization, and compare its robustness with that of pixel-space WAMs. We further evaluate Devol-ONE on Flexiv single-arm and dual-arm platforms.
\end{itemize}
\section{Related Work}
\noindent\textbf{Vision-Language-Action Models.}
Vision-Language-Action (VLA) models \citep{RT123,RT223,roboflaminggo23li,li2024towards,cogact24li,24pi0,25pi0.5,liu2024rdt,liu2025hybridvla,shukor2025smolvla,bjorck2025gr00t,wen25tinyvla,zheng2025xvla} adapt pretrained vision-language backbones to robotic control by learning $p(a_{1:H} \mid o, l)$ from demonstration trajectories. Lightweight variants further emphasize computational efficiency \citep{lin2025evo0visionlanguageactionmodelimplicit,lin2025evo1lightweightvisionlanguageactionmodel,lin2026evodepthlightweightdepthenhancedvisionlanguageaction}. Devol-ONE extends this formulation by incorporating predicted latent dynamics into action generation through layerwise joint attention.

\noindent\textbf{World-Action Models.}
World-Action Models (WAMs) use predictions of future visual states to guide action generation \citep{gr1,clover24,vpp2024hu,unified25,cheang2024gr,susie23,du2024learning,yang2023learning,du2023video,zhu2024irasim,cen2025worldvla,zhu2025unified,jang2025dreamgen,feng2025vidar,bharadhwaj2024gen2act,huang2025enerverse,zhang2025gevrm}. A recent survey organizes this space by how much of the future a method is required to generate before the action is obtained \citep{shen2026wamsurvey}. Cascaded WAMs follow an imagine-then-execute paradigm, introducing future visual observations $v_{1:T}$ as an intermediate variable:
\begin{equation}
p(a_{1:H} \mid o, l) = \int p(v_{1:T} \mid o, l)\, p(a_{1:H} \mid o, l, v_{1:T})\, dv_{1:T},
\end{equation}
first synthesizing future visual states then recovering action trajectories from these imagined outcomes via an inverse dynamics model \citep{seer24}. The predicted visual states provide future context for action selection. However, their strong dependency on future frame synthesis introduces considerable inference latency and potential error accumulation, motivating the latent alternatives discussed next.

\noindent\textbf{Implicit World Modeling and Latent Action Representations.}
A growing body of work retains the benefits of dynamics-aware policy learning while avoiding explicit pixel-level future synthesis, encoding predicted futures as compact latent representations rather than rendered frames \citep{yuan2026fastwamworldactionmodels,huang2025ladi,vlajepa2026}. One line augments the VLA objective with auxiliary future-prediction targets, such as multimodal world knowledge \citep{zhang2025dreamvla}, aligned latent features of future observations \citep{zheng2025flare}, joint understanding-and-prediction pretraining \citep{upvla25}, or visual and motion chain-of-thought reasoning \citep{cotvla25,zheng2024tracevla,zhong2025flowvla,zhong2026acot}. A parallel line learns action-relevant latent dynamics directly from large-scale action-free video via latent action models \citep{23lapo,edwards2019imitating,ye2024lapa,moto24,chen2024igor,bu2025univla,chen2025villa,yang2025como,liu2025stamo,bi2025motus,gaoa24daworld,zhang2026clapcontrastivelatentaction,Alexander2025Object,Alexander2026VLM,Alexander2025LALRS,garrido2026learning,zhang2025latent,bu2025laof}, avoiding paired action labels. Our approach focuses on coupling latent dynamics prediction and action generation within the policy through layerwise joint attention.
\section{Method}
\label{sec:method}

\begin{figure}[t]
    \centering
    \includegraphics[width=1\linewidth]{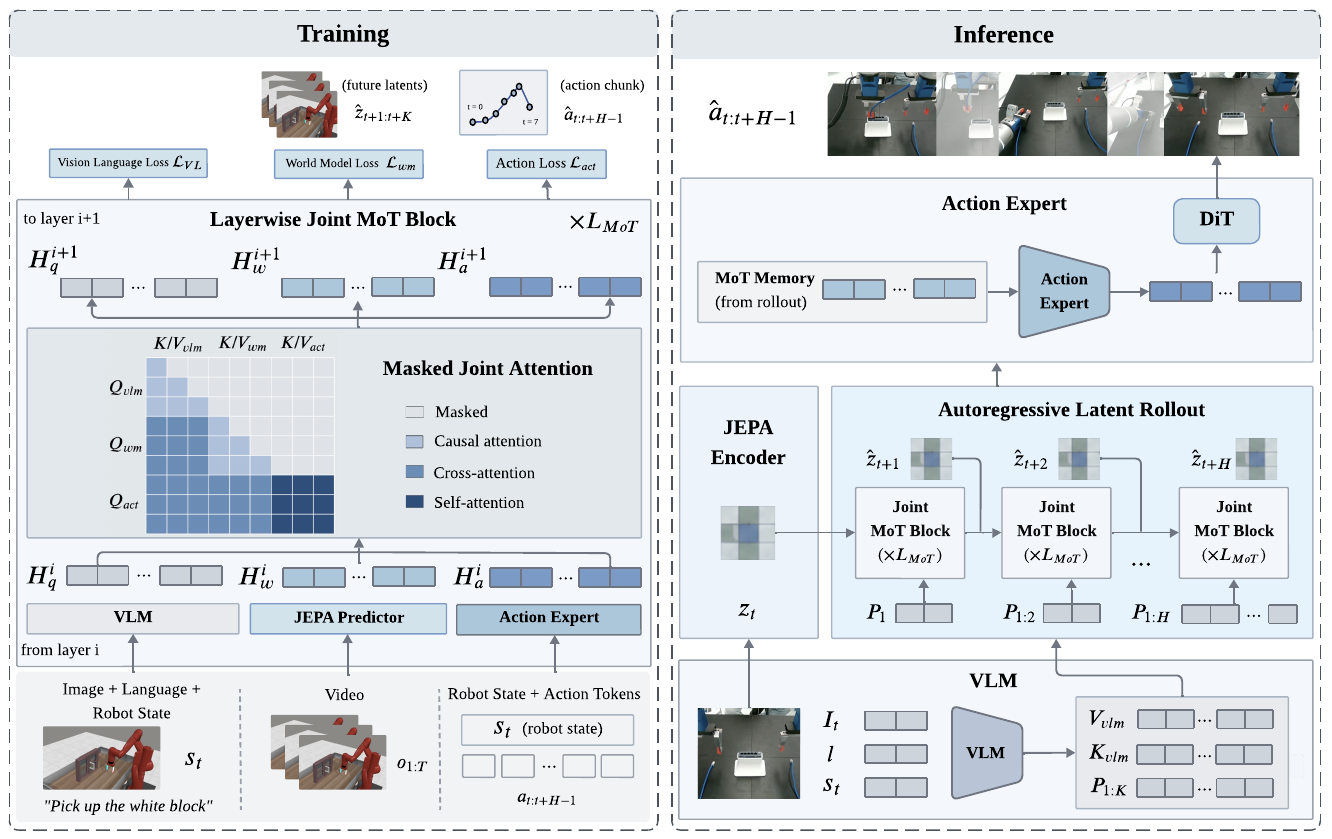}
       \caption{\textbf{Overview of Devol-ONE pipeline.} A VLM, World Model (WM), and Action Expert are trained as unified layerwise Mixture-of-Transformers blocks with masked joint attention, optimized with a world-model loss $\mathcal{L}_{\mathrm{dyn}}$ and an action loss $\mathcal{L}_{\mathrm{act}}$ (\textbf{left}). At inference (\textbf{right}), a JEPA encoder embeds the current observation, and the predictor is autoregressively rolled out through the shared MoT block to forecast future latents conditioned on the VLM. The resulting latent trajectory conditions the Action Expert, which uses a DiT head to generate the action chunk.}
    \label{fig:framework}
\end{figure}

\textbf{Overview.}
Fig. \ref{fig:framework} shows our model overall architecture.
We formulate embodied manipulation as joint conditional generation over future latent dynamics and action chunks. At time step \(t\), the policy receives observation \(o_t=(I_t,c_t,S_t)\), where \(I_t\) is the RGB image, \(c_t\) is the language instruction, and \(S_t\) is the proprioceptive state. Devol-ONE learns two objectives as shown in below.

The action policy generates a continuous action chunk \(A_t=(a_t,\dots,a_{t+H-1})\) of horizon \(H\) from the current observation:
\begin{equation}
p(A_t \mid o_t).
\end{equation}

The JEPA world model learns an autoregressive predictor over V-JEPA \citep{bardes2023vjepa1,assran2025vjepa2} latents. Given latent history \(Z_{t-L:t}=(z_{t-L},\dots,z_t)\) of \(L{+}1\) steps encoded from past observations, the predictor is trained to predict one step ahead, i.e., \((z_{t-L+1},\dots,z_{t+1})\). At inference, it is rolled out for \(K\) steps using its own predictions as history to forecast future latents \(Z_{t+1:t+K}=(z_{t+1},\dots,z_{t+K})\):
\begin{equation}
p(Z_{t+1:t+K} \mid Z_{t-L:t}, o_t).
\end{equation}
Both objectives are jointly optimized, coupling action generation with latent dynamics prediction.

\textbf{Layerwise Joint Mixture-of-Transformers with Directed Causal Attention.}
Let $X^{\mathrm{VL}}$, $X^{\mathrm{D}}$, and $X^{\mathrm{A}}$ denote the token sequences of the vision-language stream, the latent dynamics stream, and the action stream, respectively. Following the MoT formulation, every non-embedding parameter is decoupled by stream, and each stream $s\in\{\mathrm{VL},\mathrm{D},\mathrm{A}\}$ maintains its own layer normalization, query/key/value projections, and feed-forward network. At layer $\ell$, the projections are computed as
\begin{equation}
Q_s^{(\ell)}=\bar{X}_s^{(\ell)}W_{Q,s}^{(\ell)},\qquad
K_s^{(\ell)}=\bar{X}_s^{(\ell)}W_{K,s}^{(\ell)},\qquad
V_s^{(\ell)}=\bar{X}_s^{(\ell)}W_{V,s}^{(\ell)},
\end{equation}
where $\bar{X}_s^{(\ell)}=\operatorname{LN}_s^{(\ell)}(X_s^{(\ell)})$, and $d$ denotes the per-head dimension used for scaling below.

The three streams interact through a fixed directed dependency: the vision-language stream conditions the dynamics stream, and both condition the action stream. Later streams access earlier ones, whereas earlier streams do not access later ones. Specifically, the VL stream performs causal self-attention only, the dynamics stream attends to the concatenation of VL and dynamics tokens, and the action stream attends to the concatenation of VL, dynamics, and action tokens:
\begin{align}
\tilde{X}^{(\mathrm{VL})}
&=
\operatorname{Softmax}\!\left(
\frac{Q^{(\mathrm{VL})}K^{(\mathrm{VL})\top}}{\sqrt{d}}
+
M^{(\mathrm{VL})}
\right)
V^{(\mathrm{VL})},\\
\tilde{X}^{(\mathrm{D})}
&=
\operatorname{Softmax}\!\left(
\frac{Q^{(\mathrm{D})}
[K^{(\mathrm{VL})};K^{(\mathrm{D})}]^{\top}}{\sqrt{d}}
+
M^{(\mathrm{D})}
\right)
[V^{(\mathrm{VL})};V^{(\mathrm{D})}],\\
\tilde{X}^{(\mathrm{A})}
&=
\operatorname{Softmax}\!\left(
\frac{Q^{(\mathrm{A})}
[K^{(\mathrm{VL})};K^{(\mathrm{D})};K^{(\mathrm{A})}]^{\top}}{\sqrt{d}}
+
M^{(\mathrm{A})}
\right)
[V^{(\mathrm{VL})};V^{(\mathrm{D})};V^{(\mathrm{A})}],
\end{align}
where $[;]$ denotes concatenation along the token dimension, and the layer index $\ell$ is omitted for clarity. Masks \(M^{(\mathrm{VL})}\) and \(M^{(\mathrm{D})}\) are causal and enforce temporal ordering, while \(M^{(\mathrm{A})}\) enforces the directed stream dependency; attention within the action stream is bidirectional over the action chunk. In cross-stream attention, an earlier-stream key is visible only when its temporal index does not exceed that of the query. Thus, VL accesses only its causal history, dynamics accesses the VL context and its own history, and action accesses the VL context, dynamics context, and its own action history. No stream accesses a later stream or a future rollout step.

The language instruction specifies what task the model should perform. Since the dynamics stream attends to VL keys and values at every layer, this conditioning is applied at each rollout step and helps keep the predicted trajectory aligned with the task. Each sublayer is applied residually within its own stream,
\begin{equation}
\hat{X}_s^{(\ell)}
=
X_s^{(\ell)}
+
\tilde{X}_s^{(\ell)},
\qquad
X_s^{(\ell+1)}
=
\hat{X}_s^{(\ell)}
+
\operatorname{FFN}_s^{(\ell)}
\!\left(
\operatorname{LN}_s^{(\ell)}
\!\left(
\hat{X}_s^{(\ell)}
\right)
\right).
\end{equation}
With stream-specific normalization, projection, and FFN parameters, and cross-stream communication mediated solely by the directed causal attention above, the model preserves specialization of vision-language reasoning, latent dynamics, and action generation while coupling them through shared layerwise computation.

\textbf{Training time.}
The three streams are trained jointly with stream-specific objectives. The VL stream performs autoregressive next-token prediction over text tokens conditioned on the image with a causal mask, optimized by cross-entropy:
\begin{equation}
\mathcal{L}_{\mathrm{VL}}
=
-\sum_{i\in \text{text}}
\log p_\theta
\left(
x_i^{\mathrm{VL}}
\mid
x_{<i}^{\mathrm{VL}}, I_t
\right).
\end{equation}
The JEPA dynamics stream performs causal latent autoregressive prediction. Given the current latent \(z_t\) and its causal history, it predicts future latents \(\hat{z}_{t+1},\dots,\hat{z}_{t+K}\) and is trained with an L1 loss against targets from a frozen V-JEPA encoder \(E\):
\begin{equation}
\mathcal{L}_{\mathrm{dyn}}
=
\sum_{k=1}^{K}
\left\|
\hat{z}_{t+k}
-
E(I_{t+k})
\right\|_1 ,
\end{equation}
The action expert is a Diffusion Transformer (DiT) \citep{peebles23dit} with bidirectional self-attention over action tokens. It denoises a noisy action chunk \(A_t^{\tau}\) conditioned on the VL representations, predicted latent rollout, and proprioceptive state, trained with a denoising objective:
\begin{equation}
\mathcal{L}_{\mathrm{act}}
=
\mathbb{E}_{\epsilon,\tau}
\left[
\left\|
\epsilon
-
\epsilon_\theta
\left(
A_t^{\tau},\tau,
X^{\mathrm{VL}},
\hat{Z}_{t+1:t+K},
S_t
\right)
\right\|_2^2
\right],
\end{equation}
where \(A_t^{\tau}\) is the noisy action chunk at diffusion step \(\tau\). The overall training objective combines the three losses:
\begin{equation}
\mathcal{L}
=
\mathcal{L}_{\mathrm{VL}}
+
\lambda_{\mathrm{dyn}}\mathcal{L}_{\mathrm{dyn}}
+
\lambda_{\mathrm{act}}\mathcal{L}_{\mathrm{act}} ,
\end{equation}
with weights \(\lambda_{\mathrm{dyn}}\) and \(\lambda_{\mathrm{act}}\) reported in the Appendix.

\textbf{Autoregressive Latent Rollout Training.}
The dynamics targets above come from encoded future frames, whereas at inference the model has only its own predictions. Under teacher forcing, the predictor never sees the latent distribution it encounters in closed loop, and the action expert inherits this mismatch through its conditioning on the rollout. ALRT addresses this gap by unrolling the dynamics stream during training as at inference. Training begins with a teacher-forced warm-up, in which each step receives the encoded target latent, and then switches to a free-running mode, in which the predictor consumes its own output at every step:
\begin{equation}
\tilde{z}_{t+k}
=
\begin{cases}
E(I_{t+k}), & \text{warm-up},\\[2pt]
\hat{z}_{t+k}, & \text{free-running}.
\end{cases}
\end{equation}
Both \(\mathcal{L}_{\mathrm{dyn}}\) and \(\mathcal{L}_{\mathrm{act}}\) are evaluated on the self-generated trajectory after the switch. The transition is discrete rather than a sampling schedule \citep{bengio2015scheduledsamplingsequenceprediction}; thus training and inference rollouts match in the free-running phase.

\textbf{Inference time.}
At inference, the dynamics stream is rolled out as above, starting from the latent \(z_t\) encoded from the current observation and conditioned on \(X^{\mathrm{VL}}\). The resulting latent trajectory \(\hat{Z}_{t+1:t+K}\) conditions the action expert, which denoises an action chunk \(A_t\) through reverse diffusion. The robot executes the first \(H_{\mathrm{exec}}\leq H\) actions of the chunk and replans at the next step. Forecasting is performed in V-JEPA latent space, so inference requires neither future frame synthesis nor an inverse dynamics model.

\textbf{Optional CoT branch.}
In addition to the JEPA predictor, the vision-language backbone supports inference-time autoregressive rollout, producing chain-of-thought text alongside the action. This branch annotates each step with the current subtask and makes task stages explicit, for dataset construction and failure analysis. It is optional and remains disabled in all reported results unless stated otherwise. Since the VL key-value cache is shared by the dynamics stream and the action expert, we either prefill all language tokens or decode them from the prompt, feeding generated tokens back into the model at the next autoregressive step.

\begin{figure}  \centering  \includegraphics[width=1\linewidth]{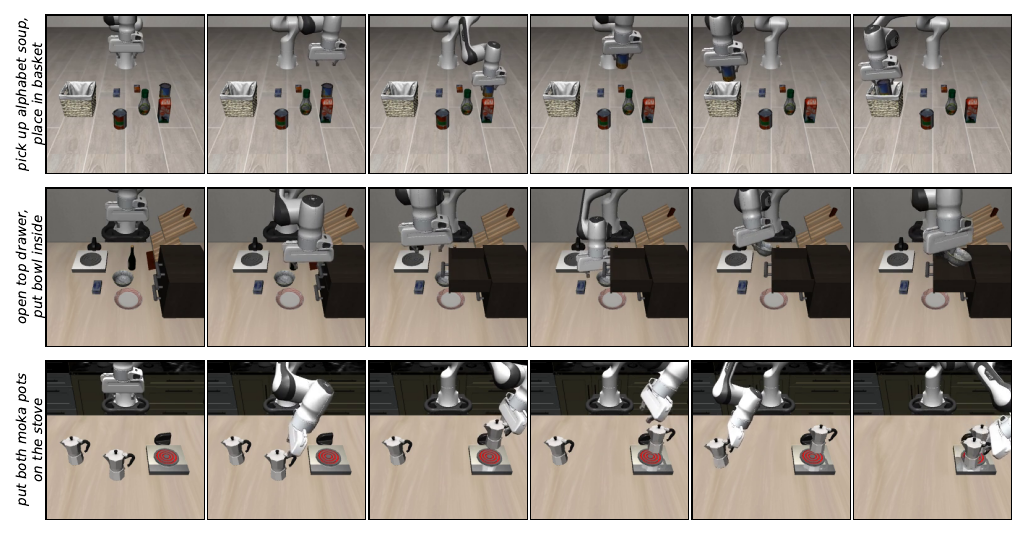}  

\caption{Example rollouts on LIBERO.}  \label{fig:libero_rollout}

\end{figure} 

\section{Experiments}
\subsection{Experimental Setup}
\textbf{LIBERO \& LIBERO-Plus.} We continue training Devol-ONE from a robot-pretrained checkpoint. We evaluate the resulting checkpoint on the standard four-suite LIBERO benchmark \citep{LIBERO23}. On LIBERO-Plus \citep{fei25libero-plus} we evaluate the checkpoint under two protocols. In the zero-shot setting, the checkpoint is evaluated directly on LIBERO-Plus without fine-tuning on its perturbed data. In the in-distribution (zero-shot) setting, we fine-tune on the LIBERO-Plus training split before evaluation. Both protocols evaluate across seven perturbation dimensions: Camera, Robot, Language, Light, Background, Noise, and Layout, each spanning the four LIBERO task suites.
\textbf{RoboTwin 2.0.} RoboTwin 2.0 \citep{mu2025robotwin} is a benchmark for bimanual manipulation with more than 50 tasks that require coordinated dual-arm control.

\textbf{Real-world benchmarks.} We evaluate single- and dual-arm manipulation on a Flexiv platform using human-teleoperated demonstrations.  Each baseline is trained separately for each task. The task definitions and evaluation procedure are described below.

Additional training and evaluation details, including training schedules and prompt configurations, are provided in Appendix~\ref{a:pretraining}.

 \subsection{Simulation Results}

\textbf{LIBERO.} Table~\ref{tab:libero_eval} compares Devol-ONE with published policies on the four LIBERO suites. Devol-ONE achieves an average success rate of 98.4\%, the highest among the compared methods. On the Long suite, it reaches 97.0\%, compared with 95.2\% for Fast-WAM, 95.8\% for VLA-JEPA, and 94.5\% for OpenVLA-OFT. It also achieves success rates of 98.4\% on Spatial, 99.6\% on Object, and 98.2\% on Goal. Figure~\ref{fig:libero_rollout} shows example rollouts.

\textbf{LIBERO-Plus.} Table~\ref{tab:libero_plus} reports zero-shot results without fine-tuning on perturbed data. Devol-ONE achieves an overall success rate of 71.4\%, the highest among the compared methods. It leads on Language, Light, and Layout, with success rates of 85.9\%, 94.5\%, and 80.5\%, respectively.

\textbf{RoboTwin 2.0.} We compare against results from the RoboTwin 2.0 leaderboard \citep{robotwin2025leaderboard}. Table~\ref{tab:robotwin} presents four representative dual-arm tasks under the Clean and Randomized settings. Devol-ONE achieves the highest Randomized success rate on all four tasks: 99.0\% on Click Alarmclock, 93.0\% on Dump Bin Bigbin, 68.5\% on Place Bread Basket, and 67.6\% on Place Can Basket. Fast-WAM achieves higher success rates under the Clean setting on three of the four tasks, but its success rates decrease to 60.0\%, 0.0\%, 0.0\%, and 0.0\%, respectively, under the Randomized setting. Per-task Clean/Randomized results are provided in Table~\ref{tab:robotwin_full_appendix}.

\begin{table}[t]
\caption{\textbf{Success rates (\%) on the LIBERO benchmark.}}
\label{tab:libero_eval}
\centering
\setlength{\tabcolsep}{8pt} 
\renewcommand{\arraystretch}{0.9} 
\setlength{\aboverulesep}{0.15ex}
\setlength{\belowrulesep}{0.15ex}
\begin{tabular*}{\textwidth}{@{\extracolsep{\fill}}l ccccc}
\toprule
\textbf{Model} & \textbf{Spatial} & \textbf{Object} & \textbf{Goal} & \textbf{Long} & \textbf{Avg} \\
\midrule
Diffusion Policy \citep{chi2023diffusion}  & 78.3  & 92.5  & 68.3  & 50.5  & 72.4  \\
Octo  \citep{team2024octo}  & 78.9  & 85.7  & 84.6  & 51.1  & 75.1  \\
OpenVLA \citep{kim2024openvla}  & 84.7  & 88.4  & 79.2  & 53.7  & 76.5  \\
SpatialVLA \citep{qu25spatialvla}  & 88.2  & 89.9  & 78.6  & 55.5  & 78.1  \\
CoT-VLA \citep{cotvla25}  & 87.5  & 91.6  & 87.6  & 69.0  & 81.1  \\
WorldVLA (512) \citep{cen2025worldvla}  & 87.6  & 96.2  & 83.4  & 60.0  & 81.8  \\
VLA-RFT \citep{li2025vlarft}  \tablefootnote{From public reports.\label{ftn:publicreports}}  & 94.4  & 94.4  & 95.4  & 80.2  & 91.1  \\
RynnVLA-002 (Disc.) \citep{cen2025rynnvla002}\hyperref[ftn:publicreports]{\textsuperscript{\getrefnumber{ftn:publicreports}}} & 94.2  & 96.8  & 94.6  & 87.6  & 93.3  \\
GR00T N1 \citep{bjorck2025gr00t}  & 94.4  & 97.6  & 93.0  & 90.6  & 93.9  \\
$\pi_0$ \citep{24pi0}  & 96.8  & 98.8  & 95.8  & 85.2  & 94.2  \\
UniVLA \citep{bu2025univla}  & 96.5  & 96.8  & 95.6  & 92.0  & 95.2  \\
$\pi_{0.5}$ \citep{25pi0.5}  & 98.8  & 98.2  & 98.0  & 92.4  & 96.9  \\
OpenVLA-OFT \citep{kim2025fine}  & 97.6  & 98.4  & 97.9  & 94.5  & 97.1  \\
VLA-JEPA \citep{vlajepa2026}  & 96.2  & 99.6  & 97.2  & 95.8  & 97.2  \\
RynnVLA-002 (Cont.) \citep{cen2025rynnvla002}  & \textbf{99.0}  & \textbf{99.8}  & 96.4  & 94.4  & 97.4  \\
Fast-WAM \citep{yuan2026fastwamworldactionmodels}  & 97.0  & 98.2  & \textbf{100.0}  & 95.2  & 97.6  \\
\midrule
\textbf{Devol-ONE (Ours)} & 98.4 & 99.6 & 98.2 & \textbf{97.0} & \textbf{98.4} \\
\bottomrule
\end{tabular*}
\end{table}
\begin{table}[H]
\caption{\textbf{Zero-shot performance on LIBERO-Plus.} All methods are trained only on the standard LIBERO dataset without fine-tuning on LIBERO-Plus dataset.}
\label{tab:libero_plus}
\centering
\small 
\setlength{\tabcolsep}{-3pt} 
\renewcommand{\arraystretch}{0.8} 
\begin{tabular*}{\textwidth}{@{\extracolsep{\fill}} l ccccccccc @{}}
\toprule
\textbf{Method} & \textbf{Camera} & \textbf{Robot} & \textbf{Language} & \textbf{Light} & \textbf{Background} & \textbf{Noise} & \textbf{Layout} & \textbf{Total} \\
\midrule
OpenVLA \citep{kim2024openvla}  & 0.8  & 3.5  & 23.0 & 8.1  & 34.8  & 15.2 & 28.5 & 15.6 \\
WorldVLA \citep{cen2025worldvla}  & 0.1  & 27.9 & 41.6 & 43.7 & 17.1  & 10.9 & 38.0 & 25.0 \\
NORA  & 2.2  & 37.0 & 65.1 & 45.7 & 58.6  & 12.8 & 62.1 & 39.0 \\
UniVLA \citep{bu2025univla}  & 1.8  & 46.2 & 69.6 & 69.0 & 81.0  & 21.2 & 31.9 & 42.9 \\
Fast-WAM \citep{yuan2026fastwamworldactionmodels}  & 44.5 & \textbf{68.9} & 60.7 & 53.7 & 37.7  & 16.4 & 78.2 & 51.5 \\
$\pi_0$ \citep{24pi0}  & 13.8 & 6.0  & 58.8 & 85.0 & 81.4  & \textbf{79.0} & 68.9 & 53.6 \\
OpenVLA-OFT\_w \citep{kim2025fine} & 10.4 & 38.7 & 70.5 & 76.8 & \textbf{93.6} & 49.9 & 69.9 & 55.8 \\
$\pi_0$-Fast\tablefootnote{FAST-tokenized variant.} \citep{pertsch2025fast}  & \textbf{65.1} & 21.6 & 61.0 & 73.2 & 73.2  & 74.4 & 68.8 & 61.6 \\
OpenVLA-OFT\_m \citep{kim2025fine}  & 55.6 & 21.7 & 81.0 & 92.7 & 91.0  & 78.6 & 68.7 & 67.9 \\
RIPT-VLA  & 55.2 & 31.2 & 77.6 & 88.4 & 91.6  & 73.5 & 74.2 & 68.4 \\
OpenVLA-OFT \citep{kim2025fine}  & 56.4 & 31.9 & 79.5 & 88.7 & 93.3  & 75.8 & 74.2 & 69.6 \\
\midrule
\textbf{Devol-ONE (Ours)} & 47.3 & 45.7 & \textbf{85.9} & \textbf{94.5} & 91.8 & 67.5 & \textbf{80.5} & \textbf{71.4} \\
\bottomrule
\end{tabular*}
\end{table}

\begin{table}[htb!]
\centering
\caption{Simulation results on the RoboTwin 2.0 leaderboard. We report success rates (\%) under \textit{Clean} and \textit{Randomized} settings for four representative tasks.}
\label{tab:robotwin}
\resizebox{\textwidth}{!}{%
\begin{tabular}{lcccccccc}
\toprule
& \multicolumn{2}{c}{\textbf{Click Alarmclock}} & \multicolumn{2}{c}{\textbf{Dump Bin Bigbin}} & \multicolumn{2}{c}{\textbf{Place Bread Basket}} & \multicolumn{2}{c}{\textbf{Place Can Basket}} \\
\cmidrule(lr){2-3} \cmidrule(lr){4-5} \cmidrule(lr){6-7} \cmidrule(lr){8-9}
\textbf{RoboTwin} & Clean & Randomized & Clean & Randomized & Clean & Randomized & Clean & Randomized \\
\midrule
ACT \citep{zhao23act}  & 32.0 & 4.0  & 68.0 & 1.0  & 6.0  & 0.0 & 1.0  & 0.0 \\
Diffusion Policy \citep{chi2023diffusion}  & 61.0 & 5.0  & 49.0 & 0.0  & 14.0 & 0.0 & 18.0 & 0.0 \\
RDT \citep{liu2024rdt}  & 61.0 & 12.0 & 64.0 & 32.0 & 10.0 & 2.0 & 19.0 & 6.0 \\
$\pi_0$ \citep{24pi0}  & 63.0 & 11.0 & 83.0 & 24.0 & 17.0 & 4.0 & 41.0 & 5.0 \\
DP3 \citep{ze20243d}  & 77.0 & 14.0 & 85.0 & 53.0 & 26.0 & 1.0 & 67.0 & 2.0 \\
starVLA \citep{starvla2026}  & 60.0 & 0.0  & 54.0 & 4.0  & 82.0 & 5.0 & 4.0  & 0.0 \\
$\pi_{0.5}$ \citep{25pi0.5}  & 65.0 & 50.0 & 92.0 & 84.0 & 63.0 & 60.0 & 56.0 & 10.0 \\
Abot-M0 \citep{yang2026abotm0vlafoundationmodel}  & 0.0  & 21.0 & 89.0 & 68.0 & 70.0 & 47.0 & 38.0 & 13.0 \\
Xiaomi Robotics-0 \citep{cai2026xiaomirobotics0opensourcedvisionlanguageactionmodel} & 97.0 & 58.0 & 94.0 & 49.0 & 54.0 & 23.0 & 52.0 & 7.0 \\
EventVLA \citep{yang2026eventvlaeventdrivenvisualevidence}  & 93.0 & 30.0 & 92.0 & 64.0 & 66.0 & 15.0 & 76.0 & 7.0 \\
X-WAM \citep{xwam}  & 95.0 & 61.0 & 88.0 & 40.0 & 82.0 & 40.0 & 63.0 & 15.0 \\
X-VLA \citep{zheng2025xvla}  & 62.0 & 43.0 & 83.0 & 41.0 & 85.0 & 52.0 & 70.0 & 9.0 \\
GalaxeaVLA \citep{liu2026g05autoregressivestreamrobot}  & 79.0 & 7.0  & 88.0 & 40.0 & 77.0 & 32.0 & 67.0 & 6.0 \\
AHA-WAM \citep{cai2026ahawamasynchronoushorizonadaptiveworldactionmodeling} & 98.0 & 17.0 & 90.0 & 27.0 & 54.0 & 3.0  & 61.0 & 2.0 \\
Spatial Forcing \citep{li2025spatialforcingimplicitspatial}  & 89.0 & 28.0 & 83.0 & 70.0 & 84.0 & 53.0 & \textbf{77.0} & 10.0 \\
Fast-WAM \citep{yuan2026fastwamworldactionmodels}  & \textbf{99.0} & 60.0 & 94.0 & 0.0  & \textbf{87.0} & 0.0 & 77.0 & 0.0 \\
GigaBrain-0.7 \citep{gigabrainteam2026gigabrain07scalingembodiedfoundation} & 98.0 & 96.0 & 90.0 & 66.0 & 60.0 & 68.0 & 20.0 & 22.0 \\
\midrule
\textbf{Devol-ONE (Ours)} & 98.0 & \textbf{99.0} & \textbf{95.0} & \textbf{93.0} & 59.0 & \textbf{68.5} & 58.0 & \textbf{67.6} \\
\bottomrule
\end{tabular}%
}
\end{table} 
\begin{table}[H]
  \centering
  \caption{\textbf{Flexiv manipulation tasks and demonstration counts.} Abbreviations are used in Table~\ref{tab:flexiv-rollout}. Details on the task and task criteria are listed in Appendix~\ref{a:tasks}}
  \label{tab:flexiv-tasks}
  \centering
  \small
  \setlength{\tabcolsep}{0pt}
  \renewcommand{\arraystretch}{0.6}
  \begin{tabular*}{\textwidth}{@{\extracolsep{\fill}} l l r @{}}
  \toprule
  Abbreviation & Task prompt (verbatim) & Episodes \\
  \midrule
  \textbf{Ethernet} & Insert the Ethernet connector  & 118 \\
  \textbf{2arm box} & Stack realsense boxes with both arms  & 200 \\
  \textbf{1arm box} & Stack realsense boxes with only the left arm  & 200 \\
  \textbf{Basket} & Place the snacks into a basket  & 200 \\
  \textbf{4cups} & Stack four different-colored cups  & 200 \\
  \bottomrule
  \end{tabular*}
\end{table}
\subsection{Real-World Evaluation}


We collect human-teleoperated demonstrations on a fixed Flexiv manipulation platform with two Flexiv Rizon \citep{flexiv2024rizon4} 7-DoF collaborative arms and three Intel RealSense D-series \citep{keselman2017intel} color cameras. One camera is mounted between the arms, and one is mounted on each wrist to capture multiple views of the workspace. Demonstrations are collected using a Meta Quest VR controller, with each episode starting from the same home position. Task prompts and demonstration counts are listed in Table~\ref{tab:flexiv-tasks}. We train Devol-ONE and the baseline models on these demonstrations, evaluate each model over 20 rollout episodes, and record the success rate. Table~\ref{tab:flexiv-rollout} reports the success rates. Devol-ONE achieves the highest average success rate across tasks, outperforming all baselines including $\pi_{0.5}$ and $\pi_0$, with particularly strong margins on the Basket and 4cups tasks. Further details can be referred to Appendix~\ref{a:robot_system}.

\begin{table*}[t]
\centering
\begin{minipage}[t]{0.55\textwidth}
\centering
\caption{Action expert conditioning on LIBERO.}
\label{tab:ablation_cond}
\small
\setlength{\tabcolsep}{2pt}
\renewcommand{\arraystretch}{0.9}
\begin{tabular}{lccccc}
\toprule
Conditioning & Spatial & Object & Goal & Long & Avg \\
\midrule
VL only & 96.2 & 98.4 & 96.8 & 92.5 & 96.0 \\
JEPA only & 97.1 & 98.8 & 97.5 & 94.2 & 96.9 \\
Final-layer fusion & \textbf{98.4} & 99.0 & \textbf{98.6} & 92.1 & 97.0 \\
\textbf{Layerwise (Ours)} & \textbf{98.4} & \textbf{99.6} & 98.2 & \textbf{97.0} & \textbf{98.4} \\
\bottomrule
\end{tabular}
\end{minipage}
\hfill
\begin{minipage}[t]{0.41\textwidth}
\centering
\caption{Effect of prediction horizon on LIBERO.}
\label{tab:ablation_frames}
\small
\setlength{\tabcolsep}{4pt}
\renewcommand{\arraystretch}{0.9}
\begin{tabular}{cccc}
\toprule
Steps $N$ & Frames & Long & Avg \\
\midrule
1 & 4 & 94.5 & 96.8 \\
2 & 6 & 96.2 & 97.5 \\
\textbf{3} & \textbf{8} & \textbf{97.0} & \textbf{98.4} \\
\bottomrule
\end{tabular}
\end{minipage}
\end{table*}

\begin{table}[htb!]
  \centering
  \caption{\textbf{Real-world rollout success rates (\%).} Full task prompts are listed in Table~\ref{tab:flexiv-tasks}.}
  \label{tab:flexiv-rollout}
  \centering
  \small
  \setlength{\tabcolsep}{4pt}
  \setlength{\rotheadsize}{3.2cm}
  \renewcommand{\arraystretch}{0.9}
  \begin{tabular*}{\textwidth}{@{\extracolsep{\fill}} l c c c c c cr @{}}
  \toprule
  {\textbf{Task}}
  & {{\textbf{Ethernet}}} & {{\textbf{2arm box}}} & {{\textbf{1arm box}}} & {{\textbf{Basket}}} & {{\textbf{4cups}}} & {{\textbf{Avg.}}}  \\
  \midrule
  GigaBrain-0.7\cite{gigabrainteam2026gigabrain07scalingembodiedfoundation} & 0.0 & 80.0 & 90.0 & 20.0 & 70.0 & 52.0 \\
  DM0.5\cite{yu2026dm0}  & 0.0 & 40.0 & 45.0 & 25.0 & 20.0 & 26.0 \\
  Fast-WAM\cite{yuan2026fastwamworldactionmodels}  & 0.0 & 55.0 & 60.0 & 0.0 & 0.0  & 23.0 \\
  $\pi_0$\cite{24pi0}  & 40.0 & 90.0 & 100.0 & 35.0 & 10.0 & 55.0 \\
  $\pi_{0.5}$\cite{25pi0.5}  & 45.0 & 100.0 & 100.0 & 40.0 & 35.0 & 64.0 \\
  \midrule
  \textbf{Devol-ONE (Ours)} & 55.0 & 100.0 & 100.0 & 50.0 & 75.0 & \textbf{76.0} \\
  \bottomrule
  \end{tabular*}
\end{table}




\subsection{Ablation Studies}

We conduct ablation studies on LIBERO to examine three aspects of Devol-ONE for the contribution of semantic and dynamics features to action generation, effect of layerwise interaction and dynamics prediction horizon. Further details for LIBERO-PLUS are listed in Appendix.

\noindent\textbf{Action expert conditioning.}
Table~\ref{tab:ablation_cond} compares four conditioning configurations. VL only and JEPA only condition the action expert on vision-language features and predicted dynamics features, respectively. The two remaining configurations use both streams, either through final-layer fusion or through the layerwise joint attention used in Devol-ONE. It is worth noting that last-layer fusion is basically equivalent to Groot-like models \citep{lin2025evo1lightweightvisionlanguageactionmodel, vlajepa2026}, by feeding last layer of hidden features into action expert instead of uniting attentions. This comparison examines both the value of combining the two information sources and the effect of how they are integrated into action generation. The full model achieves 98.4\% average success, compared with 96.0\% for VL only and 96.9\% for JEPA only, supporting the complementary value of semantic and dynamics features. Layerwise joint attention also outperforms final-layer fusion on average (98.4\% vs.\ 97.0\%), with increment on Long (97.0\% vs.\ 92.1\%).

\noindent\textbf{Prediction horizon.}
Table~\ref{tab:ablation_frames} compares prediction horizons of four, six, and eight frames. Average success increases from 96.8\% to 97.5\% and 98.4\%, while Long-suite success rises from 94.5\% to 96.2\% and 97.0\%, respectively. These results suggest that longer forecasts benefit extended manipulation sequences, supporting the eight-frame horizon used in our configuration.

\section{Conclusion}
In this paper, we present Devol-ONE, a Mixture-of-Transformers architecture that integrates vision-language understanding, latent dynamics prediction, and action generation through layerwise joint attention. By forecasting future states in V-JEPA latent space, Devol-ONE provides predictive context for action generation without synthesizing future video frames. We further introduced Autoregressive Latent Rollout Training (ALRT) to expose the policy to model-generated latent predictions during training. Experiments on LIBERO, LIBERO-Plus, RoboTwin 2.0, and real-world manipulation tasks demonstrate the effectiveness of the framework, while ablations support the benefits of complementary semantic and dynamics features and layerwise conditioning. These results highlight the potential of integrating latent world modeling into embodied policies.

\subsection*{AI use statement}
 
In this work, we used generative AI tools to assist with writing, coding, and reproducibility-related engineering support. We have not used generative AI tools for generating synthetic datasets, developing theoretical models or conceptual frameworks, formulating mathematical claims, providing critical ingredients for proving mathematical claims, assisting in the writing of proofs, proposing or refining hypotheses, designing or providing feedback on research methodology or experiments, cleaning and reformatting datasets, supporting qualitative and thematic data analysis, or interpreting results; none of these tasks were AI-assisted. Assistance with translation is not applicable to this work.
 
Specifically, we used generative AI tools to draft and edit parts of the manuscript text and improve its readability, and to assist in writing, refactoring, and debugging code used in our experiments and reproducibility efforts. All AI-assisted text was reviewed and edited by the authors, and all AI-assisted code was reviewed and tested by the authors for correctness before use. We take responsibility for the final content of this work, including text, claims, and artifacts produced with the aid of generative AI.
 




\bibliography{iclr2027_conference}
\bibliographystyle{iclr2027_conference}

\appendix
\section{Appendix}

\subsection{RoboTwin 2.0 per-task success rates}
\label{a:robotwinsuccessrate}
Table~\ref{tab:robotwin_full_appendix} reports per-task Clean/Randomized success rates on RoboTwin 2.0. We compare Devol-ONE with $\pi_{0.5}$ \citep{25pi0.5}, X-VLA \citep{zheng2025xvla}, Fast-WAM \citep{yuan2026fastwamworldactionmodels}, GigaBrain-0.7 \citep{gigabrainteam2026gigabrain07scalingembodiedfoundation}, and X-WAM \citep{xwam}. 
\subsection{RoboTwin 2.0 per-task success rates}
\label{a:robotwinsuccessrate}
Table~\ref{tab:robotwin_full_appendix} reports per-task Clean/Randomized success rates on RoboTwin 2.0. We compare Devol-ONE with $\pi_{0.5}$ \citep{25pi0.5}, X-VLA \citep{zheng2025xvla}, Fast-WAM \citep{yuan2026fastwamworldactionmodels}, GigaBrain-0.7 \citep{gigabrainteam2026gigabrain07scalingembodiedfoundation}, and X-WAM \citep{xwam}.

\begin{table*}[t]
\centering
\caption{Per-task success rates on all 50 RoboTwin 2.0 tasks (Clean/Randomized, \%). Baselines are co-trained on all 50 tasks and taken from the official RoboTwin 2.0 leaderboard; Devol-ONE is trained with single-task fine-tuning. In each column, the best result is in \textbf{bold} and the second best is \underline{underlined}. \textit{Overall} is the mean of the Clean and Randomized averages.}
\label{tab:robotwin_full_appendix}
\resizebox{\textwidth}{!}{%
\begin{tabular}{lcccccc}
\toprule
\textbf{Task} & $\pi_{0.5}$ & X-VLA & Fast-WAM & GigaBrain-0.7 & X-WAM & \textbf{Devol-ONE (Ours)} \\
\midrule
Adjust Bottle             & \textbf{99}/\underline{77} & \textbf{99}/33 & \textbf{99}/0 & \underline{98}/\textbf{98} & 92/13 & \underline{98}/\textbf{98} \\
Beat Block Hammer         & \underline{93}/23 & 87/9 & 82/0 & \textbf{94}/\underline{74} & 92/21 & 86/\textbf{93} \\
Blocks Ranking RGB        & 72/44 & 3/4 & \underline{81}/0 & 68/\textbf{76} & \textbf{85}/36 & 55/\underline{64} \\
Blocks Ranking Size       & 45/21 & \textbf{56}/25 & \underline{52}/0 & 42/\underline{28} & 33/0 & 26/\textbf{39} \\
Click Alarmclock          & 65/50 & 62/43 & \textbf{99}/60 & \underline{98}/\underline{96} & 95/61 & \underline{98}/\textbf{99} \\
Click Bell                & 31/35 & \textbf{100}/63 & 95/7 & \underline{98}/\underline{84} & \textbf{100}/52 & \textbf{100}/\textbf{100} \\
Dump Bin Bigbin           & 92/\underline{84} & 83/41 & \underline{94}/0 & 90/66 & 88/40 & \textbf{95}/\textbf{93} \\
Grab Roller               & \underline{99}/\underline{97} & \textbf{100}/20 & \textbf{100}/2 & \textbf{100}/\textbf{100} & 95/16 & \textbf{100}/\textbf{100} \\
Handover Block            & 30/12 & \textbf{77}/1.3 & \underline{66}/0 & 16/\underline{32} & 21/21 & 19/\textbf{39} \\
Handover Mic              & \underline{98}/6 & \underline{98}/1 & \underline{98}/0 & \textbf{100}/\textbf{100} & 97/8 & \underline{98}/\underline{87} \\
Hanging Mug               & 17/14 & \underline{19}/2 & 14/0 & \textbf{26}/\textbf{36} & 17/0 & 13/\underline{25} \\
Lift Pot                  & \textbf{98}/35 & \underline{96}/15 & 91/0 & 86/\underline{86} & 77/0 & \underline{96}/\textbf{98} \\
Move Can Pot              & 60/10 & 58/0 & \textbf{96}/0 & \underline{86}/\textbf{82} & 77/34 & 65/\underline{72} \\
Move Pillbottle Pad       & 46/32 & 66/3 & \textbf{92}/0 & 32/\textbf{58} & \underline{78}/26 & 43/\underline{39} \\
Move Playingcard Away     & \underline{84}/\underline{65} & \underline{84}/5 & \textbf{94}/0 & 58/\textbf{82} & 83/11 & 83/\textbf{82} \\
Move Stapler Pad          & 20/18 & \textbf{42}/0 & \underline{33}/0 & 26/\textbf{40} & 26/2 & \underline{33}/\underline{29} \\
Open Laptop               & \underline{93}/68 & 85/16 & 84/0 & 78/\underline{88} & 72/6 & \textbf{96}/\textbf{98} \\
Open Microwave            & \textbf{86}/\textbf{59} & 43/5 & 17/0 & \underline{82}/\underline{50} & 9/24 & 39/42 \\
Pick Diverse Bottles      & 66/42 & 50/30 & \underline{72}/0 & 38/\underline{50} & \textbf{74}/26 & 56/\textbf{62} \\
Pick Dual Bottles         & 72/\textbf{57} & 41/51 & \underline{87}/1 & 26/\underline{56} & \textbf{99}/55 & 34/48 \\
Place A2B Left            & \textbf{71}/\underline{49} & 53/15 & 69/0 & 68/\textbf{72} & \underline{70}/18 & 48/44 \\
Place A2B Right           & \textbf{68}/\underline{39} & 49/13 & \underline{60}/1 & 58/\textbf{68} & 59/21 & 16/11 \\
Place Bread Basket        & 63/60 & \underline{85}/52 & \textbf{87}/0 & 60/\textbf{68} & 82/40 & 59/\underline{67} \\
Place Bread Skillet       & 76/47 & \textbf{89}/17 & \underline{85}/0 & 80/\underline{50} & 82/37 & 59/\textbf{71} \\
Place Burger Fries        & 82/\underline{86} & 90/35 & \textbf{97}/0 & 82/\underline{86} & \underline{93}/36 & 85/\textbf{93} \\
Place Can Basket          & 56/10 & \underline{70}/9 & \textbf{77}/0 & 20/\underline{22} & 63/15 & 58/\textbf{67} \\
Place Cans Plasticbox     & 31/\textbf{73} & \underline{93}/\underline{60} & \textbf{98}/0 & 8/18 & \underline{93}/36 & 37/43 \\
Place Container Plate     & 92/72 & 94/39 & \textbf{100}/0 & 94/\textbf{94} & \underline{95}/52 & \underline{95}/\underline{89} \\
Place Dual Shoes          & \textbf{70}/45 & 30/12 & 67/0 & \underline{68}/\textbf{72} & 38/7 & 58/\underline{54} \\
Place Empty Cup           & \textbf{96}/80 & 90/16 & 90/0 & 90/\textbf{90} & \underline{93}/36 & 84/\underline{88} \\
Place Fan                 & \underline{66}/26 & 63/1 & 59/0 & \textbf{78}/\textbf{74} & 47/6 & 37/\underline{48} \\
Place Mouse Pad           & 34/\underline{31} & \underline{53}/0 & \textbf{55}/0 & 26/\textbf{44} & 49/6 & 0/29 \\
Place Object Basket       & 70/29 & \underline{81}/6 & \textbf{82}/0 & 74/\textbf{64} & 52/20 & 59/\underline{45} \\
Place Object Scale        & \underline{78}/\underline{38} & 76/6 & \textbf{83}/0 & 64/\textbf{54} & 66/14 & 35/36 \\
Place Object Stand        & 85/\underline{47} & \underline{91}/43 & \textbf{94}/0 & 88/\textbf{76} & 83/28 & 18/32 \\
Place Phone Stand         & 65/37 & \underline{84}/5 & \textbf{85}/0 & 36/\textbf{52} & 35/4 & 35/\underline{45} \\
Place Shoe                & 82/50 & 83/39 & \underline{87}/0 & \textbf{88}/\textbf{92} & 76/31 & 76/\underline{91} \\
Press Stapler             & 79/63 & \underline{87}/73 & \textbf{90}/3 & \textbf{90}/\underline{74} & 85/20 & \textbf{90}/\textbf{81} \\
Put Bottles Dustbin       & \textbf{78}/\underline{50} & 47/20 & 55/0 & 68/\textbf{54} & \underline{71}/21 & 33/43 \\
Put Object Cabinet        & 39/22 & 41/21 & \underline{42}/0 & \textbf{78}/\textbf{82} & 28/16 & 15/\underline{23} \\
Rotate QRcode             & \textbf{90}/21 & \underline{81}/1 & \underline{81}/0 & 54/\underline{58} & 47/2 & 79/\textbf{64} \\
Scan Object               & 49/40 & 57/9 & \textbf{68}/0 & 32/\underline{46} & 32/14 & \underline{58}/\textbf{69} \\
Shake Bottle Horizontally & \underline{99}/\textbf{100} & \textbf{100}/78 & \textbf{100}/9 & \textbf{100}/92 & 97/66 & 90/\underline{93} \\
Shake Bottle              & \textbf{100}/\textbf{100} & \textbf{100}/76 & \textbf{100}/10 & \textbf{100}/\underline{98} & \underline{96}/67 & 94/96 \\
Stack Blocks Three        & \underline{77}/25 & 0/0 & \textbf{89}/0 & 60/\textbf{78} & 72/\underline{26} & 0/10 \\
Stack Blocks Two          & 91/43 & 10/2 & \textbf{99}/0 & 80/\textbf{92} & \underline{97}/\underline{76} & 44/50 \\
Stack Bowls Three         & \textbf{81}/52 & 77/9 & \underline{80}/0 & 70/\textbf{64} & 69/38 & 53/\underline{60} \\
Stack Bowls Two           & \textbf{94}/70 & 85/15 & \textbf{94}/0 & 90/\underline{74} & \underline{93}/67 & 82/\textbf{88} \\
Stamp Seal                & \underline{53}/\underline{21} & 50/2 & 30/0 & 52/\textbf{54} & \textbf{58}/16 & 20/18 \\
Turn Switch               & \underline{52}/25 & 40/6 & 39/3 & 42/\underline{52} & \textbf{68}/4 & 51/\textbf{53} \\
\midrule
\textbf{Average}          & \underline{70.7}/46.0 & 68.0/20.9 & \textbf{77.8}/1.9 & 66.8/\textbf{67.9} & 70.0/25.8 & 58.0/\underline{62.2} \\
\textbf{Overall}          & 58.3 & 44.5 & 39.8 & \textbf{67.4} & 47.9 & \underline{60.1} \\
\bottomrule
\end{tabular}%
}
\end{table*}

\paragraph{Overall comparison.}
Averaged over all 50 tasks, Devol-ONE achieves 58.0\% under the Clean setting and 62.2\% under the Randomized setting, for an overall average of 60.1\%. This places it second among the compared methods, behind GigaBrain-0.7 (67.4\%) and ahead of $\pi_{0.5}$ (58.3\%), X-WAM (47.9\%), X-VLA (44.5\%), and Fast-WAM (39.8\%). The advantage comes mainly from the Randomized setting, where Devol-ONE's average is second only to GigaBrain-0.7 (67.9\%) and exceeds $\pi_{0.5}$ (46.0\%), X-WAM (25.8\%), X-VLA (20.9\%), and Fast-WAM (1.9\%). At the task level, Devol-ONE obtains the best Randomized score on 20 of the 50 tasks and the second best on another 17, placing it in the top two on 37 tasks; in 16 of the 17 tasks where it ranks second, it trails only GigaBrain-0.7. In head-to-head comparisons under the Randomized setting, Devol-ONE outperforms Fast-WAM, X-WAM, X-VLA, and $\pi_{0.5}$ on 50, 46, 46, and 37 tasks, respectively.

\paragraph{Sensitivity to domain randomization.}
The clearest difference between methods is the gap between the two settings. Most baselines degrade substantially under domain randomization: the average Clean-to-Randomized drop is 24.7 points for $\pi_{0.5}$, 44.1 for X-WAM, and 47.0 for X-VLA, while Fast-WAM, despite the highest Clean average (77.8\%), falls to near-zero Randomized success on almost every task. Devol-ONE shows no such degradation; its Randomized average is 4.1 points above its Clean average, a pattern otherwise observed only for GigaBrain-0.7 (1.1 points). 
We note that Devol-ONE scores higher under Randomized than Clean on 34 of the 50 tasks. Because the Randomized setting is designed to be strictly harder, we do not interpret this as a benefit of randomization itself, and leave a closer analysis to future work.

\paragraph{Limitations.}
Devol-ONE has the lowest Clean average among the compared methods, with the best Clean score on 5 tasks and the second best on 7. Its failures concentrate on tasks that require precise placement relative to a target or fine-grained reasoning over multiple objects, such as Place Mouse Pad (0/29), Stack Blocks Three (0/10), Place A2B Right (16/11), Hanging Mug (13/25), Put Object Cabinet (15/23), Place Object Stand (18/32), and Stamp Seal (20/18). Finally, Devol-ONE is trained with single-task fine-tuning, whereas all baselines are co-trained on the full 50-task suite, so the per-task comparison is not a strictly controlled one.

\begin{figure}  \centering  \includegraphics[width=1\linewidth]{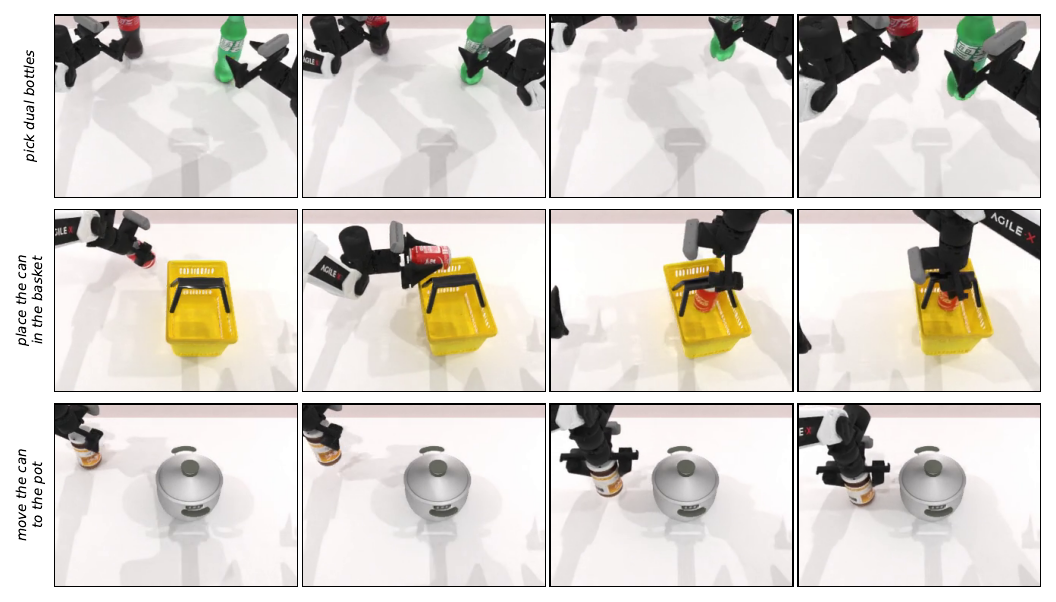}  \caption{Example rollouts on RoboTwin 2.0.}  \label{fig:robotwin_rollout}
\end{figure}
%

\subsection{Additional LIBERO-Plus Results}
\label{app:liberoplus}

We report two additional evaluations on LIBERO-Plus~\citep{fei25libero-plus}: in-distribution
fine-tuning and the action-expert conditioning ablation under the zero-shot protocol. All
evaluations use the full test set of 10{,}030 episodes.

\subsubsection{In-Distribution Fine-Tuning}
\label{app:liberoplus_sft}

We fine-tune Devol-ONE on the LIBERO-Plus training split and evaluate after 50k and 150k steps
(Table~\ref{tab:liberoplus_sft}). Fine-tuning raises the total success rate from 71.4\% in the
zero-shot setting to 82.2\% and 84.4\%, with the largest gains on Camera and Noise. Longer
training improves every factor, most notably Robot ($+$5.1) and Language ($+$3.1).

\begin{table}[h]
\centering
\caption{\textbf{In-distribution results on LIBERO-Plus (\%).} $\Delta$: change from 50k to 150k steps. \#Ep.: evaluation episodes
per factor. Robot denotes robot initial-state perturbations.}
\label{tab:liberoplus_sft}
\small
\begin{tabular}{lrcccr}
\toprule
Factor     & \#Ep.      & Zero-shot & SFT 50k & SFT 150k & $\Delta$ \\
\midrule
Camera     & 1{,}599    & 47.3      & 92.1    & 93.9     & $+$1.8 \\
Robot      & 1{,}550    & 45.7      & 43.1    & 48.2     & $+$5.1 \\
Language   & 1{,}537    & 85.9      & 81.7    & 84.8     & $+$3.1 \\
Light      & 1{,}142    & 94.5      & 97.2    & 97.9     & $+$0.7 \\
Background & 1{,}076    & 91.8      & 96.6    & 96.8     & $+$0.2 \\
Noise      & 1{,}601    & 67.5      & 92.9    & 94.3     & $+$1.4 \\
Layout     & 1{,}525    & 80.5      & 79.3    & 81.2     & $+$1.9 \\
\midrule
Total      & 10{,}030   & 71.4      & 82.2    & 84.4     & $+$2.2 \\
\bottomrule
\end{tabular}
\end{table}

\subsubsection{Action-Expert Conditioning on LIBERO-Plus}
\label{app:liberoplus_ablation}

We evaluate the conditioning variants of Table~\ref{tab:ablation_cond} on LIBERO-Plus under
the zero-shot protocol (Table~\ref{tab:liberoplus_ablation}). The layerwise model achieves the best
total (71.4\%), exceeding VL only by 2.4 points and JEPA only by 6.3 points, and leads on five of
the seven factors. The two single-source variants are complementary: JEPA only is the most robust
to robot initial-state perturbations, whereas VL only is stronger on Language, Camera, and Noise.

\begin{table}[h]
\centering
\caption{\textbf{Action-expert conditioning on LIBERO-Plus (zero-shot, \%).}  Best
result per column in bold.}
\label{tab:liberoplus_ablation}
\small
\setlength{\tabcolsep}{4pt}
\begin{tabular}{lcccccccc}
\toprule
Conditioning     & Camera        & Robot         & Language      & Light         & Background    & Noise         & Layout        & Total \\
\midrule
VL only          & \textbf{48.7} & 43.1          & 82.8          & 90.6          & 90.1          & 62.1          & 79.1          & 69.0 \\
JEPA only        & 37.5          & \textbf{55.4} & 72.3          & 92.7          & 89.9          & 46.8          & 77.6          & 65.1 \\
Layerwise (Ours) & 47.3          & 45.7          & \textbf{85.9} & \textbf{94.5} & \textbf{91.8} & \textbf{67.5} & \textbf{80.5} & \textbf{71.4} \\
\bottomrule
\end{tabular}
\end{table}

\subsection{Pretraining details}
\label{a:pretraining}

\noindent\textbf{Backbones and world model.} The dynamics-stream predictor consumes the V-JEPA2 encoder's features from the current/past observations and is trained to predict the encoder's features on \emph{future} frames (Section~\ref{sec:method}); Table~\ref{tab:backboneparams} lists the corresponding sizes.

\begin{table}[H]
\centering
\caption{Backbone and world-model hyperparameters.}
\label{tab:backboneparams}
\begin{tabular}{lr}
\toprule
\textbf{Hyperparameter} & \textbf{Value} \\
\midrule
\textbf{Vision-language stream:} &  \\
Backbone & Qwen3-VL-2B-Instruct \\
Hidden size & 2048 \\
Attention implementation & SDPA \\
\midrule
\textbf{Dynamics stream:} &  \\
Encoder & V-JEPA2 ViT-L/16 \citep{assran2025vjepa2} \\
Encoder checkpoint & \texttt{vjepa2-vitl-fpc64-256} \\
Predictor depth & 12 \\
Predictor attention heads & 8 \\
Temporal window (frames) & 8 \\
Camera views & 3 \\
Latent dynamics tokens / timestep & 32 \\
\bottomrule
\end{tabular}%
\end{table}

\noindent\textbf{Action expert.} The action expert is a DiT-Base \citep{peebles23dit} flow-matching head; Table~\ref{tab:actionexperparams} lists its hyperparameters.

\begin{table}[H]
\centering
\caption{Action expert hyperparameters.}
\label{tab:actionexperparams}
\begin{tabular}{lr}
\toprule
\textbf{Hyperparameter} & \textbf{Value} \\
\midrule
Action dimensions & 14 \\
State dimensions & 16 \\
Future action window size $K$ & 6 \\
Past action window size & 0 \\
Action horizon $H$ & 7 \\
Repeated diffusion steps & 8 \\
Flow-matching time sampling: Beta $\alpha$ & 1.5 \\
Flow-matching time sampling: Beta $\beta$ & 1.0 \\
Flow-matching time sampling: rescale $s$ & 0.999 \\
Timestep buckets & 1000 \\
Inference timesteps & 4 \\
Number of target vision tokens & 32 \\
\midrule
\textbf{Diffusion Parameters:} &  \\
Cross-attention width & 2048 \\
Norm type & \texttt{ada\_norm} \\
Number of layers & 16 \\
Output dimensions & 1024 \\
Positional embeddings & none \\
Dropout & 0.2 \\
\bottomrule
\end{tabular}%
\end{table}

\noindent\textbf{Data.} The robot-action stream mixes AgiBotWorld-ReinforcementLearning archives (Home and Industry environments) with DROID \citep{khazatsky2024droid}; both sources carry frame-level sub-task annotations, used as the CoT-conditioning instruction described in Section~\ref{sec:method} rather than the coarser episode-level instruction. An auxiliary Something-Something~V2 \citep{goyal2017something} stream with no action labels is co-trained to expose the vision-language and dynamics streams to broader human motion. Table~\ref{tab:datamixture} lists the mixture composition.

\begin{table}[H]
\centering
\caption{Pretraining data mixture.}
\label{tab:datamixture}
\begin{tabular}{lr}
\toprule
\textbf{Property} & \textbf{Value} \\
\midrule
\textbf{Robot-action stream:} &  \\
AgiBotWorld-RL archives (Home + Industry) & 78 \\
AgiBotWorld-RL sample weight (per archive) & 1.0 \\
DROID sample weight & 150 \\
Composition (DROID $:$ AgiBotWorld-RL) & $\approx$65.8\% : 34.2\% \\
Action representation & delta-position, delta-rotation-vector \\
Instruction conditioning & frame-level CoT sub-task label \\
Per-device batch size & 24 \\
\midrule
\textbf{Auxiliary video stream:} &  \\
Source & Something-Something~V2 \citep{goyal2017something} \\
Action labels & none \\
Loss terms contributed & $\mathcal{L}_{\mathrm{VL}}, \mathcal{L}_{\mathrm{dyn}}$ (no $\mathcal{L}_{\mathrm{act}}$) \\
Per-device batch size & 16 \\
\midrule
\textbf{Both streams:} &  \\
Model input resolution & $224\times224$ \\
Source decode resolution & $256\times256$ \\
\bottomrule
\end{tabular}%
\end{table}

\noindent\textbf{Optimization.} Table~\ref{tab:optimparams} lists the optimizer, learning-rate schedule, distributed-training, and loss-weighting hyperparameters (Section~\ref{sec:method}).

\begin{table}[H]
\centering
\caption{Optimization hyperparameters.}
\label{tab:optimparams}
\begin{tabular}{lr}
\toprule
\textbf{Hyperparameter} & \textbf{Value} \\
\midrule
\textbf{Optimizer (AdamW):} &  \\
$\beta_1$ & 0.9 \\
$\beta_2$ & 0.95 \\
$\epsilon$ & $10^{-8}$ \\
Weight decay & $10^{-8}$ \\
\midrule
\textbf{Learning rate:} &  \\
Peak LR: dynamics predictor \& fusion params. & $3\times10^{-5}$ \\
Peak LR: pretrained VL backbone & $1\times10^{-5}$ \\
Peak LR: action expert & $1\times10^{-4}$ \\
Warmup steps & 10{,}000 (10\%) \\
Schedule & cosine decay \\
Decay floor & $10^{-6}$ \\
\midrule
\textbf{Gradients \& distributed training:} &  \\
Gradient clipping (global norm) & 1.0 \\
Gradient accumulation steps & 1 \\
GPUs & 4 \\
Effective batch size (robot-action $/$ video) & 96 $/$ 64 \\
Precision & mixed precision (bf16) \\
Memory optimization & gradient checkpointing, DeepSpeed ZeRO-2 \\
\midrule
\textbf{Loss weights:} &  \\
Dynamics loss $\lambda_{\mathrm{dyn}}$ & 0.1 \\
CoT text loss $\lambda_{\mathrm{VL}}$ & 0.1 \\
Action loss $\lambda_{\mathrm{act}}$ (implicit) & 1.0 \\
\midrule
Checkpoint interval (steps) & 10{,}000 \\
\bottomrule
\end{tabular}%
\end{table}

\subsection{Details of real-world rollout}
\label{a:robot_system}
 The robot rig used for teleoperation is a fixed, two-arm Flexiv manipulation cell consisting of two Flexiv Rizon \citep{flexiv2024rizon4} 7-DoF collaborative arms. It is controlled through Flexiv's RDK \texttt{flexivrdk} with Python bindings. The left arm is a Flexiv Rizon~4 arm, and the right arm is Flexiv Rizon~4R arm, a variant with force and torque sensing. The right arm's force and torque sensing abilities are not used for this experiment. The platform is positioned over a black tabletop and illuminated with artificial light. Natural light is blocked to reduce illumination changes across recording sessions.

Each arm uses a WRobot gripper, model \texttt{W-EPGC-50-150-B-C-NM-L5}. Each gripper is controlled through direct register writes using Modbus RTU over a USB-to-serial adapter at \SI{115200}{\baud}. The interface uses registers for target position, speed, and grip force, which are initialized upon connection. The interface represents the gripper state as a binary variable with hysteresis: it is set to the closed state at $<\SI{30}{\percent}$ open and the open state at $>\SI{90}{\percent}$ open.

Three Intel RealSense~D-series \citep{keselman2017intel} color cameras are used, with a $640 \times 480$ pixel resolution at \SI{30}{\Hz} frame rate. One camera is mounted between the two arms, and one is mounted on each wrist. The cameras are driven with \texttt{pyrealsense2}. The cameras return \texttt{bgr8} color frames without depth data. The frames are JPEG-encoded with a quality setting of 90. The cameras are calibrated using a checkerboard/ChArUco board with $6 \times 9$ squares, a square size of \SI{30}{\mm}, and a marker size of \SI{20}{\mm}. 

\subsubsection{Tasks and Task Criteria}
\label{a:tasks}

\begin{figure}[H]
  \centering
  \begin{tabular}{cc}
    \textbf{First frame} & \textbf{Final frame} \\[2pt]
    \includegraphics[width=0.47\linewidth]{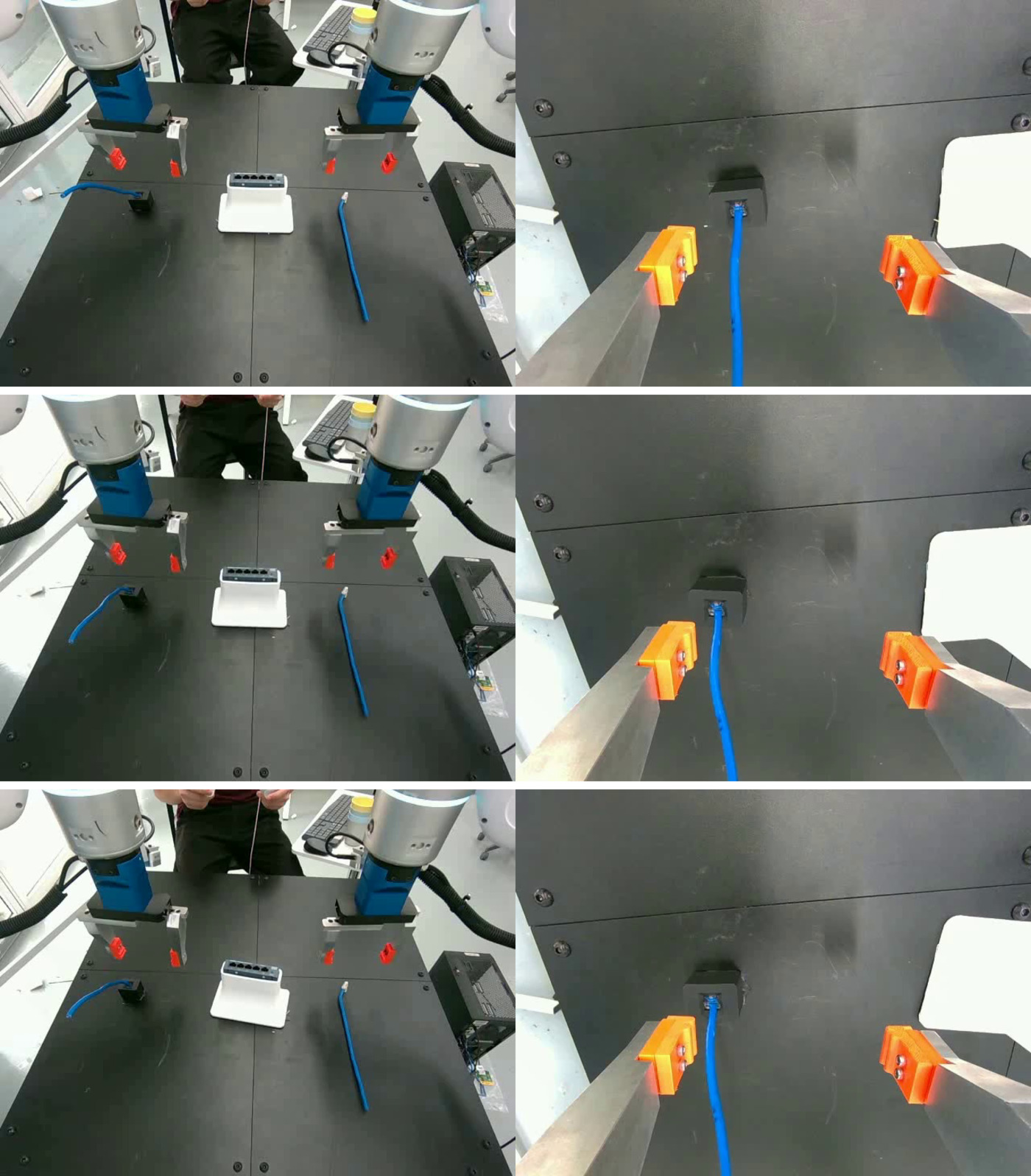} &
    \includegraphics[width=0.47\linewidth]{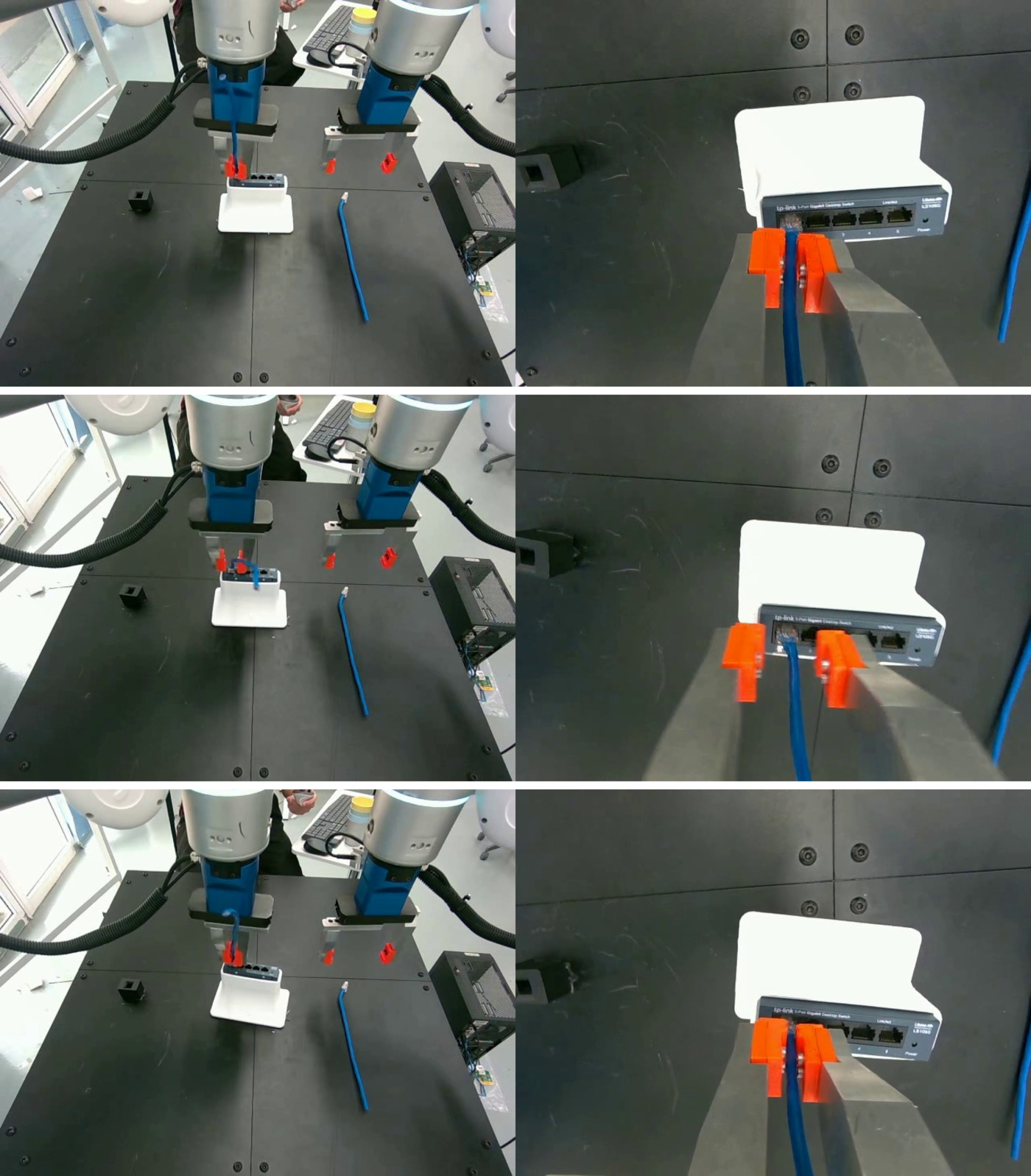} \\
  \end{tabular}
  \caption{Three sample episodes of the \textbf{Ethernet} task (rows, top to bottom), each showing the egocentric view and the left-wrist view at the first and final frame.}
  \label{fig:task_ethernet}
\end{figure}

\begin{figure}[H]
  \centering
  \begin{tabular}{cc}
    \textbf{First frame} & \textbf{Final frame} \\[2pt]
    \includegraphics[width=0.47\linewidth]{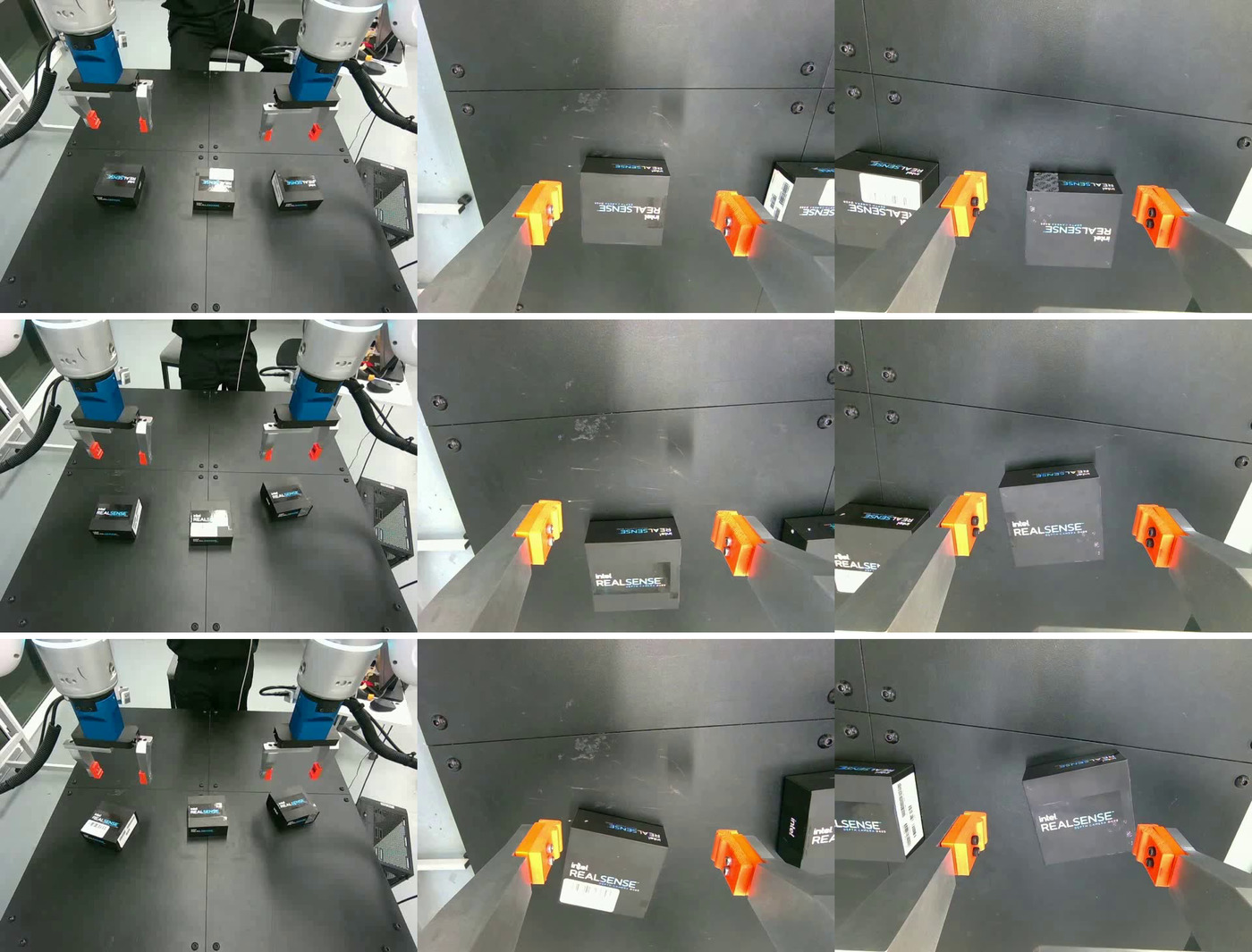} &
    \includegraphics[width=0.47\linewidth]{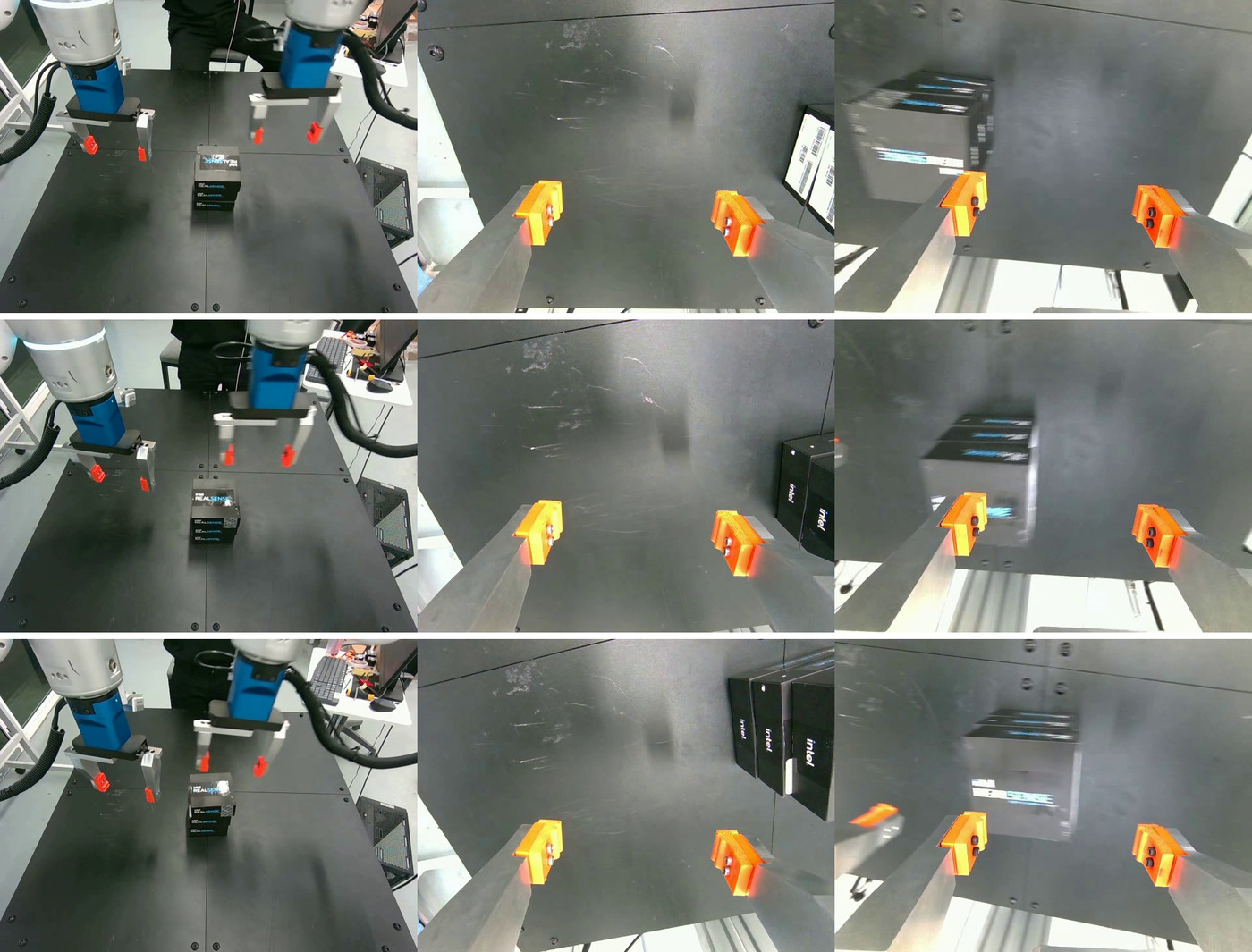} \\
  \end{tabular}
  \caption{Three sample episodes of the \textbf{2arm box} task variant (dual-arm; rows, top to bottom), each showing the egocentric view and both wrist views at the first and final frame.}
  \label{fig:task_2arm_box}
\end{figure}

\begin{figure}[H]
  \centering
  \begin{tabular}{cc}
    \textbf{First frame} & \textbf{Final frame} \\[2pt]
    \includegraphics[width=0.47\linewidth]{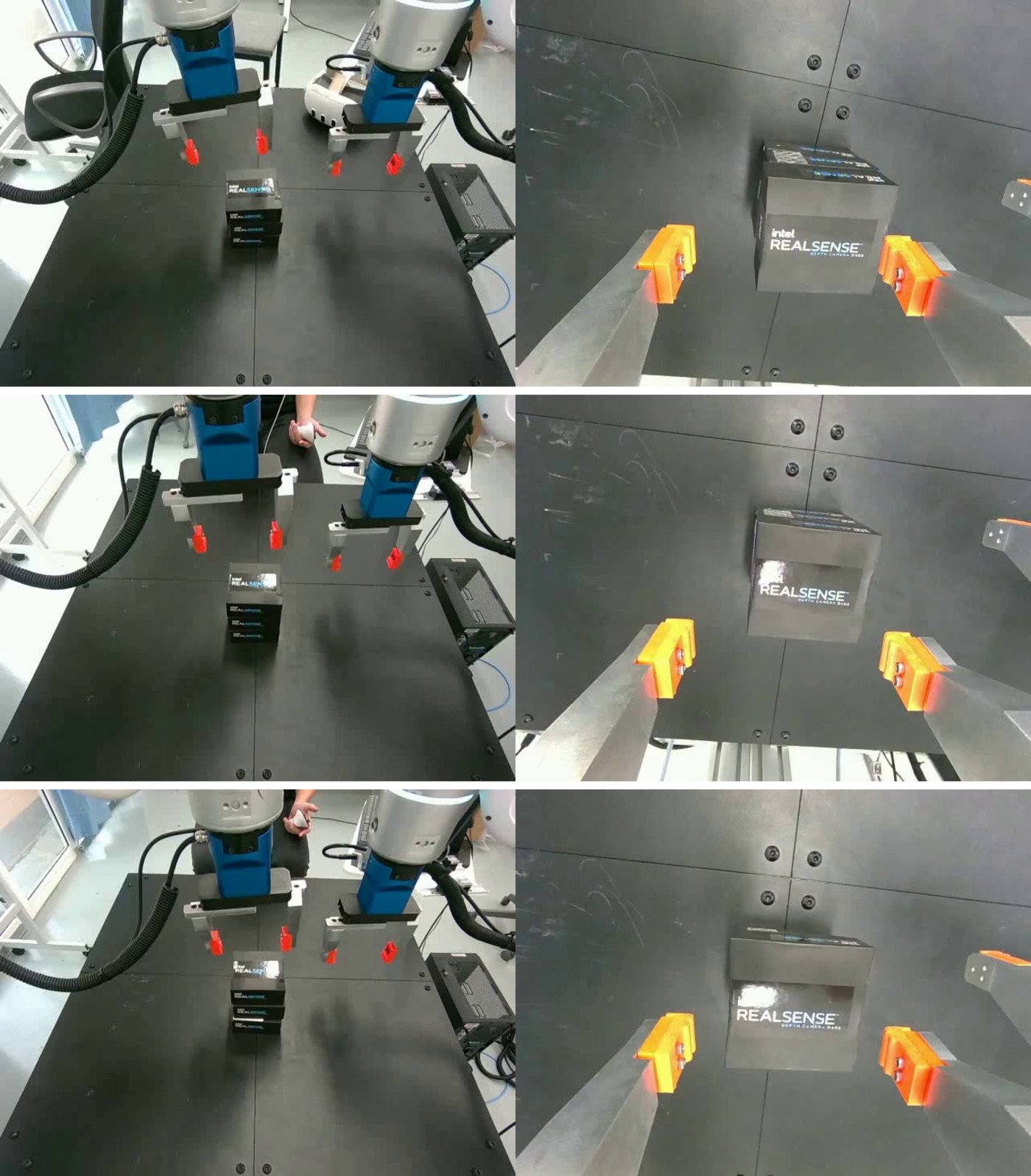} &
    \includegraphics[width=0.47\linewidth]{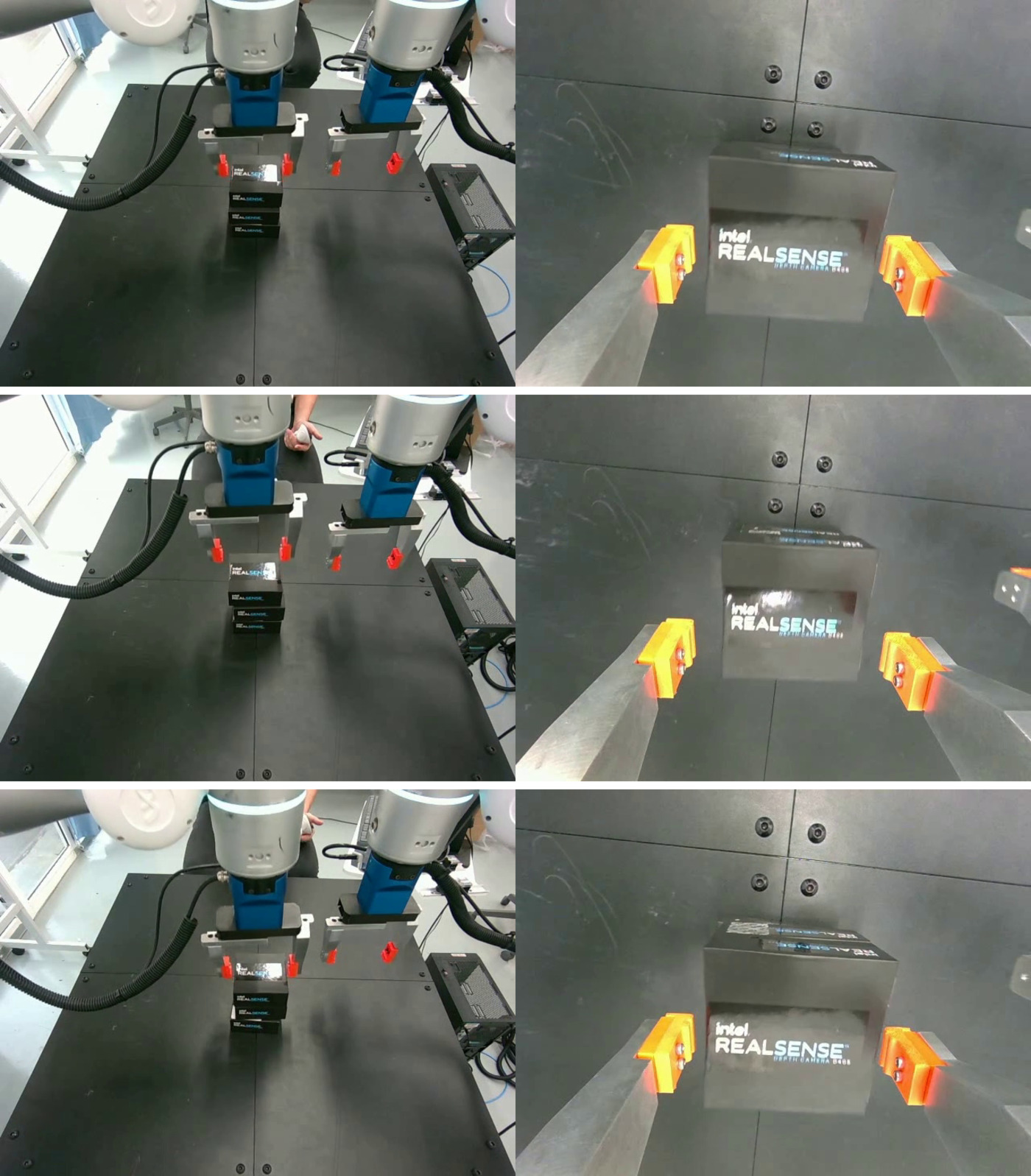} \\
  \end{tabular}
  \caption{Three sample episodes of the \textbf{1arm box} task variant (single, left arm only; rows, top to bottom).}
  \label{fig:task_1arm_box}
\end{figure}

\begin{figure}[H]
  \centering
  \begin{tabular}{cc}
    \textbf{First frame} & \textbf{Final frame} \\[2pt]
    \includegraphics[width=0.47\linewidth]{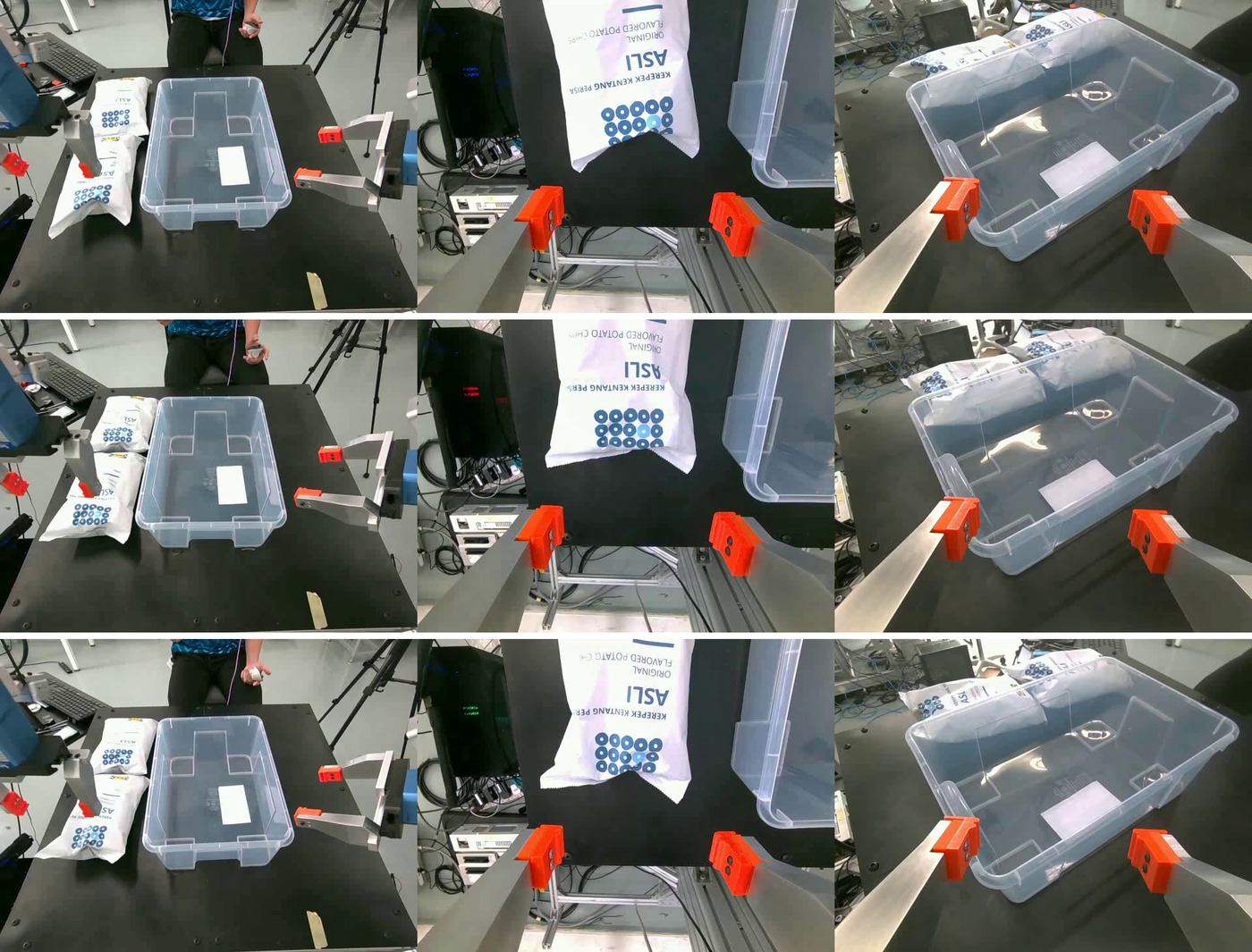} &
    \includegraphics[width=0.47\linewidth]{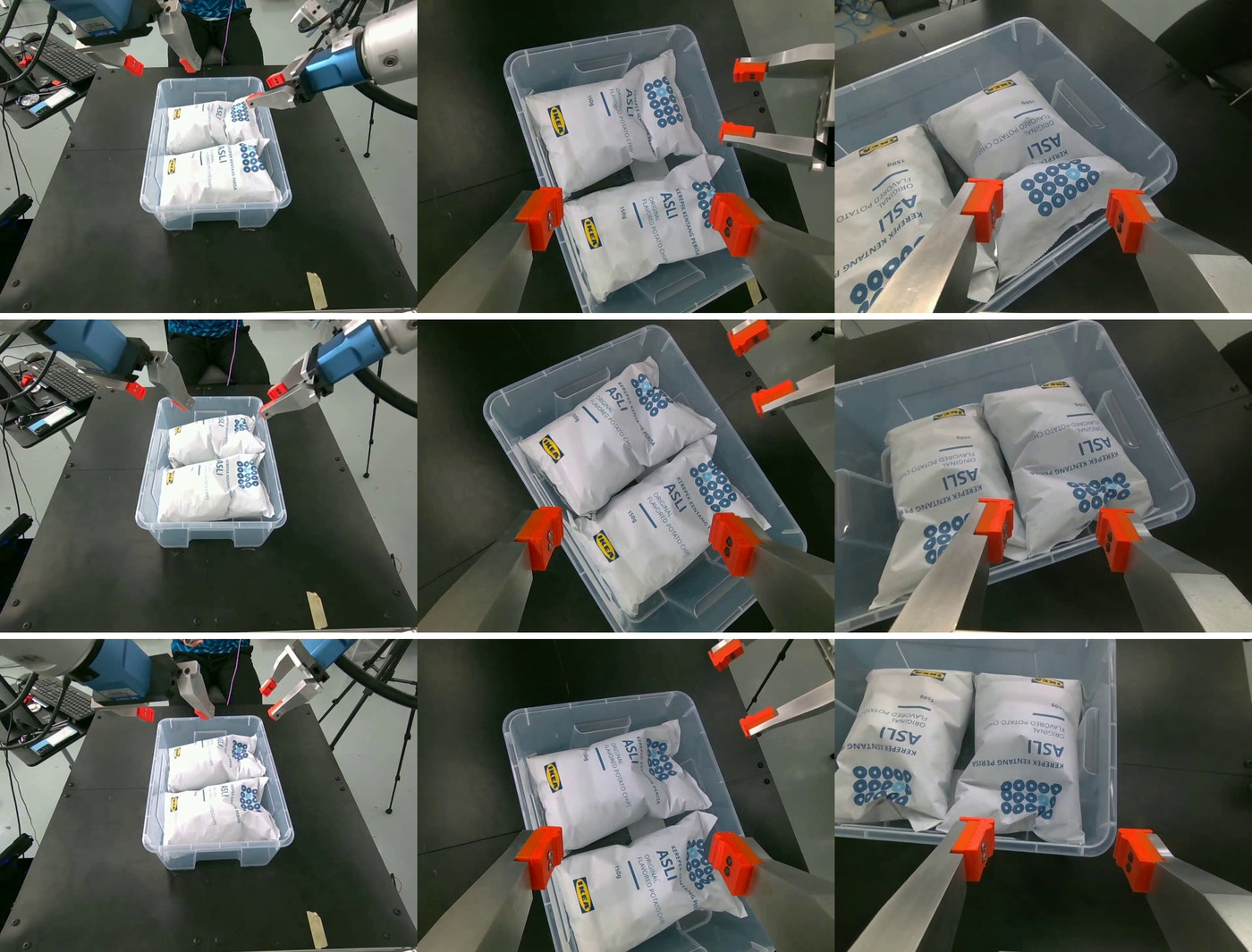} \\
  \end{tabular}
  \caption{Three sample episodes of the \textbf{Basket} task (rows, top to bottom).}
  \label{fig:task_basket}
\end{figure}

\begin{figure}[H]
  \centering
  \begin{tabular}{cc}
    \textbf{First frame} & \textbf{Final frame} \\[2pt]
    \includegraphics[width=0.47\linewidth]{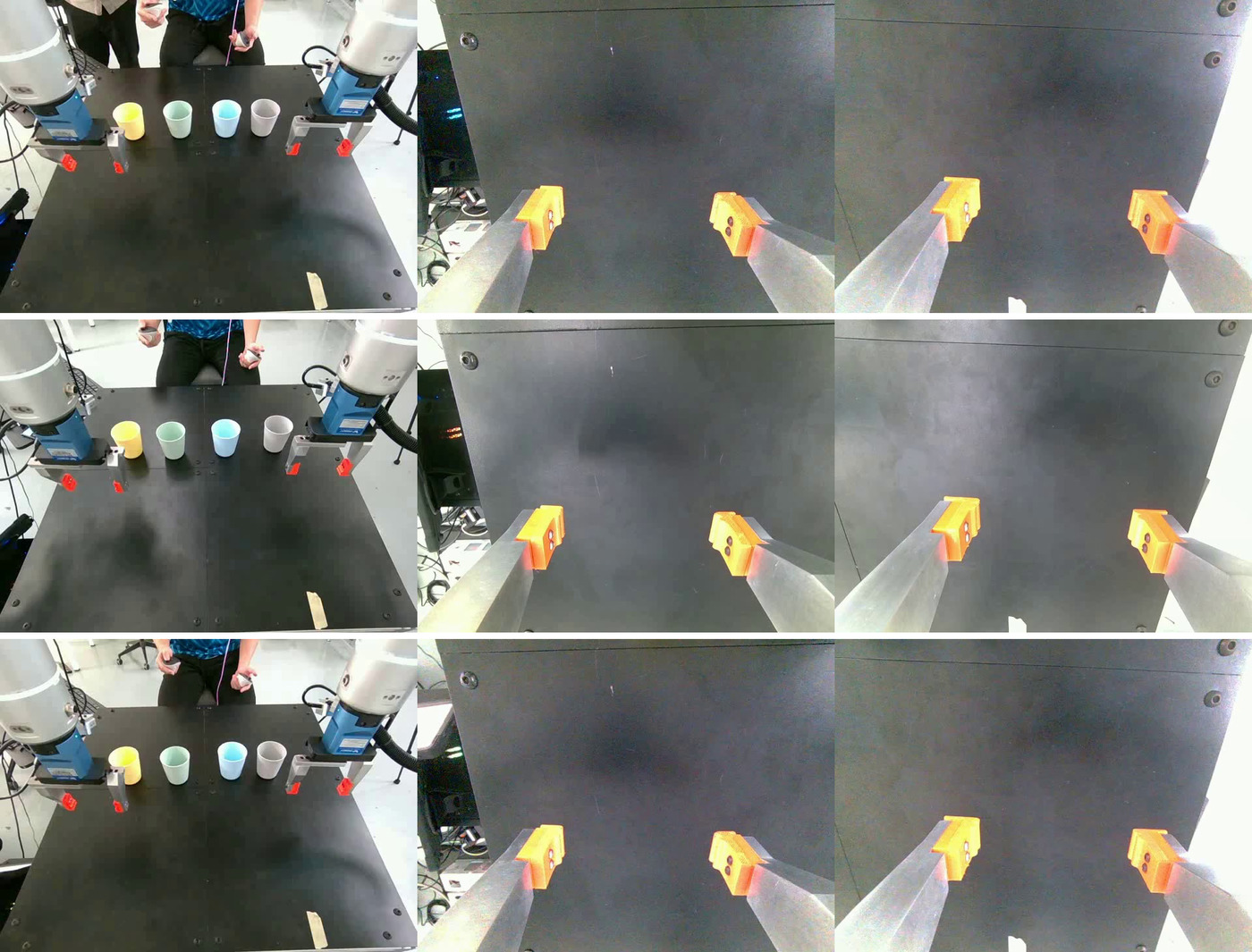} &
    \includegraphics[width=0.47\linewidth]{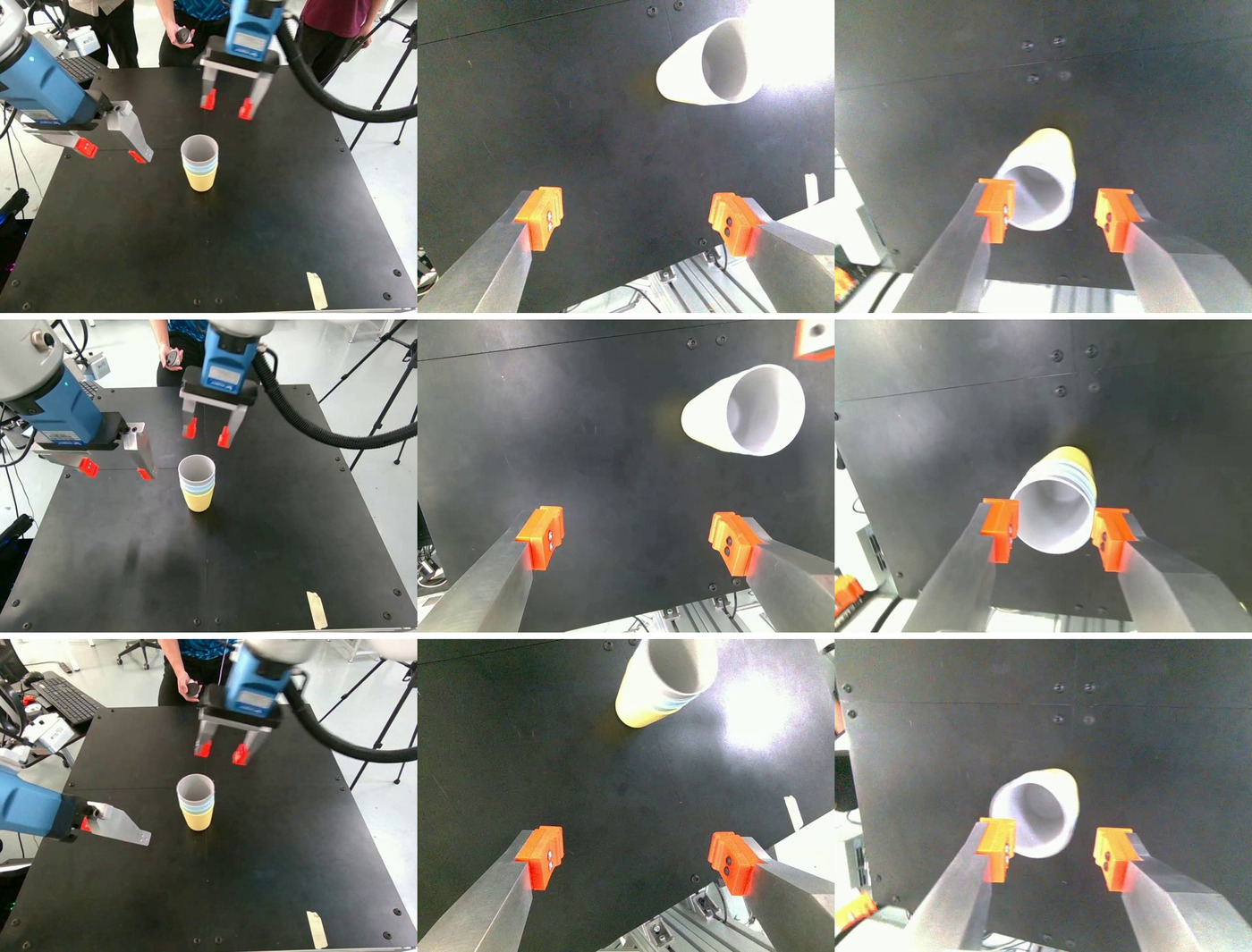} \\
  \end{tabular}
  \caption{Three sample episodes of the \textbf{4cups} task (rows, top to bottom).}
  \label{fig:task_4cups}
\end{figure}

The tasks detailed in Table~\ref{tab:flexiv-tasks} are as follows. Figures~\ref{fig:task_ethernet}--\ref{fig:task_4cups} show, for three sample episodes per task, the first and final frame of the egocentric view together with the corresponding wrist-camera view(s).

\textbf{``Insert the Ethernet Connector''} - This task is performed with the left arm only. A network switch with Ethernet receptacles is present at roughly the center of the table. Two Ethernet cables are placed on the table to the left and right of the switch; each cable has one end already secured in a fixture and the other end loose. The robot must successfully pick up the loose end of a cable and fully plug it into a receptacle on the switch. This is done twice, serially, with a single (left) arm.

\textbf{``Stack realsense [\textit{sic}] boxes with both arms/with only the left arm''} - Three Intel RealSense boxes are scattered around the table, and the robot must successfully stack them on top of one another, either using both arms in coordination (\textbf{2arm box}) or using only the left arm (\textbf{1arm box}).

\textbf{``Place the snacks into a basket''} - Two snack bags are placed on the table next to an open plastic basket, and the robot must place the snacks into the basket using both arms. The task is considered successful if at least one of the snack bags is successfully placed in the bag.

\textbf{``Stack four different-colored cups''} - Four different-colored cups are placed on the table, and the robot must stack them in no particular order, using both arms. The task is considered successful if at least three cups are stacked.

\end{document}